%% file: log_2022.tex
\documentclass{article}
 \usepackage[eatrack]{log_2023}		

\usepackage{booktabs}            
\usepackage{multirow}            
\usepackage{amsfonts}            
\usepackage{graphicx}
\usepackage{duckuments}          
\usepackage{float}
\usepackage{caption}
\usepackage{subcaption}
\usepackage{arydshln}
\usepackage{algorithm}
\usepackage{algpseudocode}
\usepackage{siunitx}
\usepackage{rotating}

\usepackage[numbers,compress,sort]{natbib}

\newcommand{\plan}[1]{}

\definecolor{light-gray}{gray}{0.95}
\newcommand{\code}[1]{\colorbox{light-gray}{\small \texttt{#1}}}
\title[SALSA-CLRS]{SALSA-CLRS: A Sparse and Scalable Benchmark for Algorithmic Reasoning}

\author[J. Minder et al.]{%
Julian Minder\\
ETH Zurich\\
\email{jminder@ethz.ch}
\And
Florian Grötschla\thanks{Equal contribution.}\\
ETH Zurich\\
\email{fgroetschla@ethz.ch}
\And
Joël Mathys\footnotemark[1]\\
ETH Zurich\\
\email{jmathys@ethz.ch}
\And
Roger Wattenhofer\\
ETH Zurich\\
\email{wattenhofer@ethz.ch}
}

\begin{document}

\maketitle

\begin{abstract}
    We introduce an extension to the CLRS algorithmic learning benchmark, prioritizing scalability and the utilization of sparse representations.
    Many algorithms in CLRS require global memory or information exchange, mirrored in its execution model, which constructs fully connected (\textit{not sparse}) graphs based on the underlying problem.
    Despite CLRS's aim of assessing how effectively learned algorithms can generalize to larger instances, the existing execution model becomes a significant constraint due to its demanding memory requirements and runtime (\textit{hard  to scale}). However, many important algorithms do not demand a fully connected graph; these algorithms, primarily distributed in nature, align closely with the message-passing paradigm employed by Graph Neural Networks.
    Hence, we propose SALSA-CLRS, an extension of the current CLRS benchmark specifically with \textit{scalability} and \textit{sparseness} in mind. Our approach includes adapted algorithms from the original CLRS benchmark and introduces  new problems from distributed and randomized algorithms.
    Moreover, we perform a thorough empirical evaluation of our benchmark. Code is publicly available at \url{https://github.com/jkminder/salsa-clrs}.
\end{abstract}

\section{Introduction}

Neural algorithmic reasoning combines the learning power of neural networks with the principles of algorithmic thinking. This fusion aims to promote logical reasoning and the ability to extrapolate. This is widely considered a weak spot for neural methods.
Algorithms take various shapes and deal with sets, strings, images, or geometry. Several prominent and beautiful algorithms are concerned with graphs and networks. Graph algorithms usually take up a significant portion of algorithmic textbooks such as the CLRS textbook~\cite{cormen2022introduction} and the CLRS benchmark \cite{velivckovic2022clrs}, which is based on that textbook. Interestingly, the CLRS benchmark translates \textit{every} algorithmic problem into a common graph-based format. 
This approach yields the significant advantage of utilizing a single architecture across various scenarios. However, the emphasis on algorithmic diversity and unification in CLRS introduces significant constraints that hinder scalability.

The CLRS-30 dataset contains 30 algorithms operating within a centralized execution model that facilitates global information exchange, which is essential for numerous algorithms.
This global information exchange is enabled by enforcing all problems to operate on a complete graph – each node can communicate with every other node, resulting in quadratic communication costs.
To maintain information on the original topology, the CLRS framework augments these complete graphs with flags on each edge to indicate whether the edge exists in the input.
This strategy has several limitations. While CLRS highlights its proficiency in assessing \emph{out-of-distribution} (OOD) capabilities, the reliance on a fully connected graph execution model imposes significant memory and computation constraints. This challenge is particularly pronounced as \emph{graph} algorithms are often designed with sparse graphs in mind \cite{cormen2022introduction}. 

\begin{figure}
    \centering
    \includegraphics[width=0.8\columnwidth]{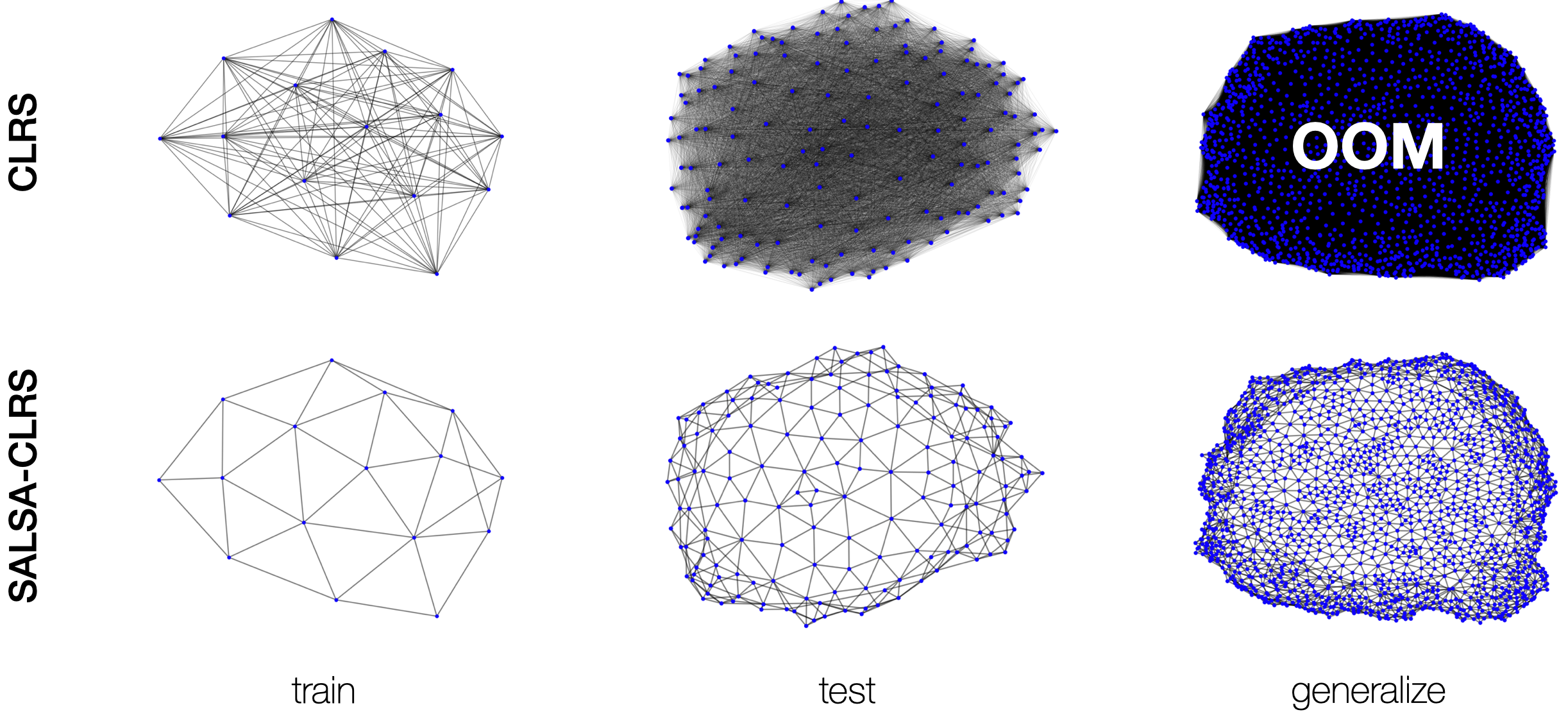}
    \caption{A visualization of the difference between the graph representation in SALSA-CLRS and CLRS. When dealing with large complete graphs, the memory demands become exceedingly impractical, leading to occurrences of \emph{Out-Of-Memory} (OOM) errors. SALSA-CLRS enables the evaluation of scalable architectures on graphs up to 100 times the size of the training graphs.}
    \label{fig:SALSA_main}
\end{figure}

Furthermore, when learning algorithms that guarantee correctness for any input size, evaluating models across a diverse range of large-scale inputs is crucial, as many studies have highlighted \cite{RecNN, tang2020towards, bansal2022end, engelmayer2023parallel, mahdavi2023towards, bevilacqua2023causal}. Apart from considering large-scale test graphs, relying solely on a single graph generation mechanism can yield false conclusions about OOD performances~\cite{mahdavi2023towards}. The CLRS library in principle allows more flexibility and a custom generation. However, the default CLRS-30 dataset used for benchmarking provides OOD test graphs, limited to only four times the size of the training graphs, and both training and test graphs stem from the same graph generation mechanism.
While under the CLRS execution model, moderately larger graphs ($10$x) might still be feasible on modern hardware, much larger graphs – in the order of $100$-fold scaling – become impossible to run due to their demanding memory requirements (Figure \ref{fig:SALSA_main}).  

To address these challenges, we propose a more concise strategy. We focus solely on \emph{graph} algorithms, which can follow a distributed execution model, thus reducing reliance on global memory and information flow. This allows a transition to a sparse execution model. Furthermore, building upon the findings presented in \cite{engelmayer2023parallel}, which underscore the superior learning and OOD performance of parallelized algorithms compared to their sequential counterparts, we also emphasize the importance of encompassing problems from the realm of distributed and randomized algorithms. Towards this end, we introduce SALSA-CLRS, a \textbf{S}parse \textbf{A}lgorithmic \textbf{L}earning benchmark for \textbf{S}calable \textbf{A}rchitectures. Extending CLRS, our benchmark \emph{i)} leverages a sparse execution mode to enable OOD test sets that cover graphs $100$ times the size of training sets, \emph{ii)} adds new graph generators for sparse and diverse graphs, thus enabling a more thorough OOD evaluation and \emph{iii)} incorporates distributed and randomized algorithms that align more closely with the execution models used by Graph Neural Networks (GNNs).

\section{SALSA-CLRS Benchmark}

The SALSA-CLRS Benchmark follows the structure of CLRS~\cite{velivckovic2022clrs}. Each data point comprises a graph with $n$ nodes and an algorithm trajectory. Each trajectory comprises a set of input, intermediate, and output features. Specifically, the input features capture the input state of the algorithm, along with positional identifiers for nodes to resolve tie-breaking scenarios. The intermediate features, referred to as hints, correspond to interim values of algorithm variables. These hints provide insight into the algorithm's inner workings and act as a means to encourage models to adhere closely to the algorithm's execution. It is worth noting that execution without hints is possible and may even be beneficial, as demonstrated in Section \ref{seq:empirical_eval}. Lastly, the output features directly relate to the solution of the given problem. Moreover, each data point contains a trajectory length, defining the number of steps required to solve the algorithm. Every feature is associated with a location – either a \emph{node}, an \emph{edge}, or the entire \emph{graph} – and possesses a corresponding type. 
SALSA-CLRS provides both pre-defined train-validation-test splits, facilitating model comparison and the capability to generate new data tailored to individual requirements. Beyond what CLRS-30 offers, SALSA-CLRS comes with diverse graph types to explore OOD capabilities further. For comprehensive information, see Appendix \ref{app:ds:stats}. The benchmark is implemented in PyG \cite{pyg} and built with extendability in mind.

\subsection{Algorithms}
SALSA-CLRS encompasses a set of six algorithms, adapting four from the original CLRS paper and introducing two novel additions from the field of distributed and randomized algorithms. The four CLRS algorithms were selected to ensure the representation of input, hint, and output features on the sparse graph: Breadth-first search (BFS), Depth-first search (DFS), Dijkstra, and Maximum Spanning Tree (MST). Please refer to Appendix \ref{app:algorithms} for more details.
While the algorithms introduced by CLRS-30 are inspired by sequential algorithms in the CLRS textbook, although in some cases heavily parallelized, the message-passing paradigm – essentially the driving mechanism behind GNNs – aligns closely with distributed computing principles. To encompass this perspective, we extend our benchmark by introducing two new distributed algorithms, drawn from \emph{Mastering Distributed Algorithms} \cite{wattenhofer2020mastering}. Numerous distributed algorithms incorporate randomness as a crucial component of their computation. In light of this, we enhance the CLRS framework by including the concept of randomness. In cases where an algorithm necessitates randomness, we precompute random values and treat them as regular input to the algorithm. We introduce two new algorithms: Distributed Maximal Independent Set (MIS) and Distributed Eccentricity. A description of both can be found in Appendix~\ref{app:algorithms}.

\subsection{Graph Types} \label{subseq:graphtypes}

Building upon investigations~\cite{mahdavi2023towards} of CLRS and different graph types, we enrich the diversity of graph types compared to CLRS-30. While CLRS-30 works exclusively on Erdös-Renyi (ER) random graphs, the study by \citet{mahdavi2023towards} underscores the limitation of relying solely on ER graphs to assess the OOD capabilities of architectures. Recognizing this, we propose that broadening the spectrum of graph types is pivotal for a more comprehensive OOD evaluation. SALSA-CLRS comes with three distinct graph generation mechanisms: Erdös-Renyi graphs (ER) \cite{ergraphs}, Watts Strogatz graphs (WS) \cite{wsgraphs} –– and Delaunay Graphs. In contrast to CLRS-30, we reduce the ER edge probability to just above the minimum to maintain graph connectivity. WS graphs belong to the category of small-world graphs and exhibit a low clustering coefficient \cite{barthelemy1999small}. 
While still sparse, WS graphs show a very different structure. Delaunay graphs are planar and hence inherently sparse. We refer to Appendix \ref{app:ds:graphs}~for associated graph parameters. 

\section{Empirical Evaluation} \label{seq:empirical_eval}

In this section, we undertake an empirical evaluation by comparing three baseline models. Our analysis involves a comparison of training scenarios with and without hints, followed by comprehensive testing across all SALSA-CLRS test sets. This evaluation sheds light on deficiencies in the models on OOD test sets and therefore affirms the importance of the SALSA-CLRS benchmark.

\paragraph{Architectures}
We use the same Encode-Process-Decode \cite{battaglia2018relational} from CLRS, but propose a slight simplification. We omit the re-encoding of decoded hints to update the node hidden states. This results in a simplification of the computational graph, making the architectures more scalable. 
We compare three baseline processors, a GRU \cite{cho2014gru} adapted GIN\footnote{Dijkstra and MST require edge weights, so we use GINE~\cite{hu2019strategies}.} module \cite{xu2018powerful}, RecGNN~\cite{RecNN}, a recurrent message-passing GNN for algorithmic tasks and PGN \cite{velivckovic2020pointer}, which has shown promising performance on the original CLRS benchmark. All architectures incorporate skip connections, implemented by forwarding the encoded input and the two most recent hidden states to the processor. This mechanism aids in mitigating vanishing gradient \cite{hochreiter1998vanishing} issues. For a comprehensive overview of the Encode-Process-Decode architecture, our proposed changes, and the baselines, please see Appendix \ref{app:eval:arch}.

\paragraph{Experiments} Each baseline model is trained for each algorithm with and without the inclusion of hints. Every run is confined to 100 epochs with early stopping. The batch size is eight graphs. All reported values are means over five runs. For more details on the metrics and the experiments, see Appendix \ref{app:eval}.

\subsection{Evaluation}
\begin{table}[]
\centering
\caption{Scores for both models on all algorithms, reported as percentages. The used metric for algorithms is denoted under the algorithm name. Models are trained only on ER graphs of size up to $n=16$ without hints (first column) and evaluated on larger sizes and different distributions (other columns).}
\scriptsize
\addtolength{\tabcolsep}{-0.7em}
\include{results/main_results}
\centering
\label{tab:main_results}
\end{table}

In Table \ref{tab:main_results}, we showcase the performance of two baseline models on all SALSA-CLRS algorithms. Note, as we increase the graph size, all models show a clear decline in performance. Furthermore, we observe significant performance disparities among different graph types. Remarkably, different algorithms show varying degrees of sensitivity to different graph types. For example, BFS shows stability when applied to larger ER graphs, but its performance drops on large Delaunay graphs. DFS shows the opposite behaviour. Similarly, the architectures show sensitivity to algorithms. For example, RecGNN shows the best extrapolation performance on BFS, while PGN is clearly the best on MIS. In general, the PGN model is often the best performer, in particular for DFS and Eccentricity, and for MIS we even see a very strong performance up to the largest graph sizes. It is worth mentioning that, consistent with previous findings~\cite{mahdavi2023towards}, the incorporation of hints does not lead to performance improvements across the board (see Tables \ref{tab:node_acc_all} and \ref{tab:graph_acc_all}). More details can be found in Appendix~\ref{app:eval}.  It is important to emphasize the pivotal role of metrics selection. An example: Despite seemingly excellent Node Accuracy scores of both baselines on BFS, the graph accuracy shows a completely different picture (see Table \ref{tab:bfs}). For larger graph instances, almost all graphs are predicted incorrectly, despite achieving a near-perfect Node Accuracy.
 These findings underscore SALSA-CLRS's effectiveness in comprehensively evaluating architectural vulnerabilities in terms of both scalability and graph diversity.

\section{Conclusion}

As traditional algorithms are invariant to input size, scalability and extrapolation are important when evaluating learned algorithmic reasoning models. Thus, we introduce SALSA-CLRS, an extension to CLRS designed for scalable architectures and sparse graph representations. Addressing the limitations of the original CLRS benchmark, SALSA-CLRS focuses on graph problems that align with distributed execution models. This orientation fosters scalability and improved assessment of generalization capabilities, particularly for larger graph instances. In addition to four CLRS problems, SALSA-CLRS incorporates two additional algorithms rooted in distributed and randomized paradigms. By including diverse \emph{out-of-distribution} (OOD) test sets, which involve graphs up to 100 times the scale of the training set and encompass various graph types, our empirical evaluation underscores the critical role of such extrapolation for a comprehensive assessment of algorithmic reasoning. These OOD tests unveil several limitations that might remain concealed when examined solely within the CLRS dataset's confines. 
 SALSA-CLRS serves as a tool for advancing Neural Algorithmic Reasoning, facilitating the evaluation of scalable architectures on sparse graphs.
\bibliographystyle{unsrtnat}
\bibliography{reference}

\newpage
\appendix
\section{Appendix}

\subsection{Related Work} \label{app:relwork}
\paragraph{Algorithmic Learning}
In recent years, the field of Algorithmic Learning has witnessed significant advancements, driven by a convergence of ideas from neural network architecture and algorithmic reasoning. To effectively tackle algorithmic tasks, models must incorporate a notion of variable computation length that enables extrapolation to handle larger input states. Consequently, proposals have emerged such as the differentiable Neural Turing machine \cite{graves2014neural} or RNNs with the capacity to generalize across varying input lengths \cite{gers2001lstm}. Schwarzschild et al. \cite{schwarzschild2021can} successfully showed the extrapolation capabilities of a recurrent architecture based on CNNs \cite{lecun1998gradient} with residual connections on a series of algorithmic tasks, including mazes, prefix sums, and chess problems.  

Recently, graph-based methods have gained traction, because of their capability to model different sizes of inputs. In particular, the use of Graph Neural Networks (GNNs) has become a focus for algorithmic learning. Notable instances of this trend include applications to problems like SAT, TSP, and shortest path computations through algorithmic alignment \cite{selsam2018learning, velivckovic2021neural, palm2018recurrent, joshi2020learning}. Theoretical investigations have established links between GNNs and dynamic programming algorithms, along with the parallel computing paradigm \cite{dudzik2022graph, loukas2019graph, engelmayer2023parallel}. Recent efforts have turned to improving the extrapolation capabilities on extending extrapolation capabilities to handle larger graph instances in the context of algorithmic reasoning problems \cite{tang2020towards, xu2020neural}. Grötschla et al. \cite{RecNN} present architectures capable of scaling up to sizes 1000 times that of the training data.

\paragraph{CLRS Benchmark}
Introducing the CLRS benchmark, Veličković et al. \cite{velivckovic2022clrs} offer a comprehensive benchmark featuring CLRS-30 a dataset with over 30 algorithms designed for algorithmic reasoning tasks. This benchmark represents algorithms as graphs with task-specific inputs, outputs, and intermediate states called hints. The CLRS has triggered a wide range of follow-up work. Ibarz et al. \cite{ibarz-generaliser} propose a generalist algorithmic learner, a single model capable of simultaneously tackling all CLRS-30 algorithms. CLRS-30 evaluations include simple \emph{Out-Of-Distribution} (OOD) tests with graphs four times the size of the training graphs. Mahdavi et al. \cite{mahdavi2023towards} provide an in-depth exploration of OOD generalization within the CLRS framework, emphasizing the need for diversified test sets consisting of more varied graphs. Several studies suggest that the inclusion of hints, as suggested by the CLRS, is not necessarily beneficial  \cite{mahdavi2023towards, bevilacqua2023causal, rodionov2023nohints}. Bevilacqua et al. \cite{bevilacqua2023causal} introduce Causal Regularization, a data augmentation technique applied to hints, which enhances OOD generalization capabilities. Their work indicates the effectiveness of hints when employed correctly. Other architectural approaches to tackling the CLRS benchmark include \cite{diao2023relational, rodionov2023nohints, georgiev2023neural}. Notably, it has been shown that parallel counterparts of the sequential algorithms implemented in CLRS prove to be more efficient to learn and execute of neural architectures, subsequently also leading to improved OOD predictions \cite{engelmayer2023parallel}.

\subsection{Algorithms} \label{app:algorithms}
\begin{figure}
    \centering
    \includegraphics[width=0.8\columnwidth]{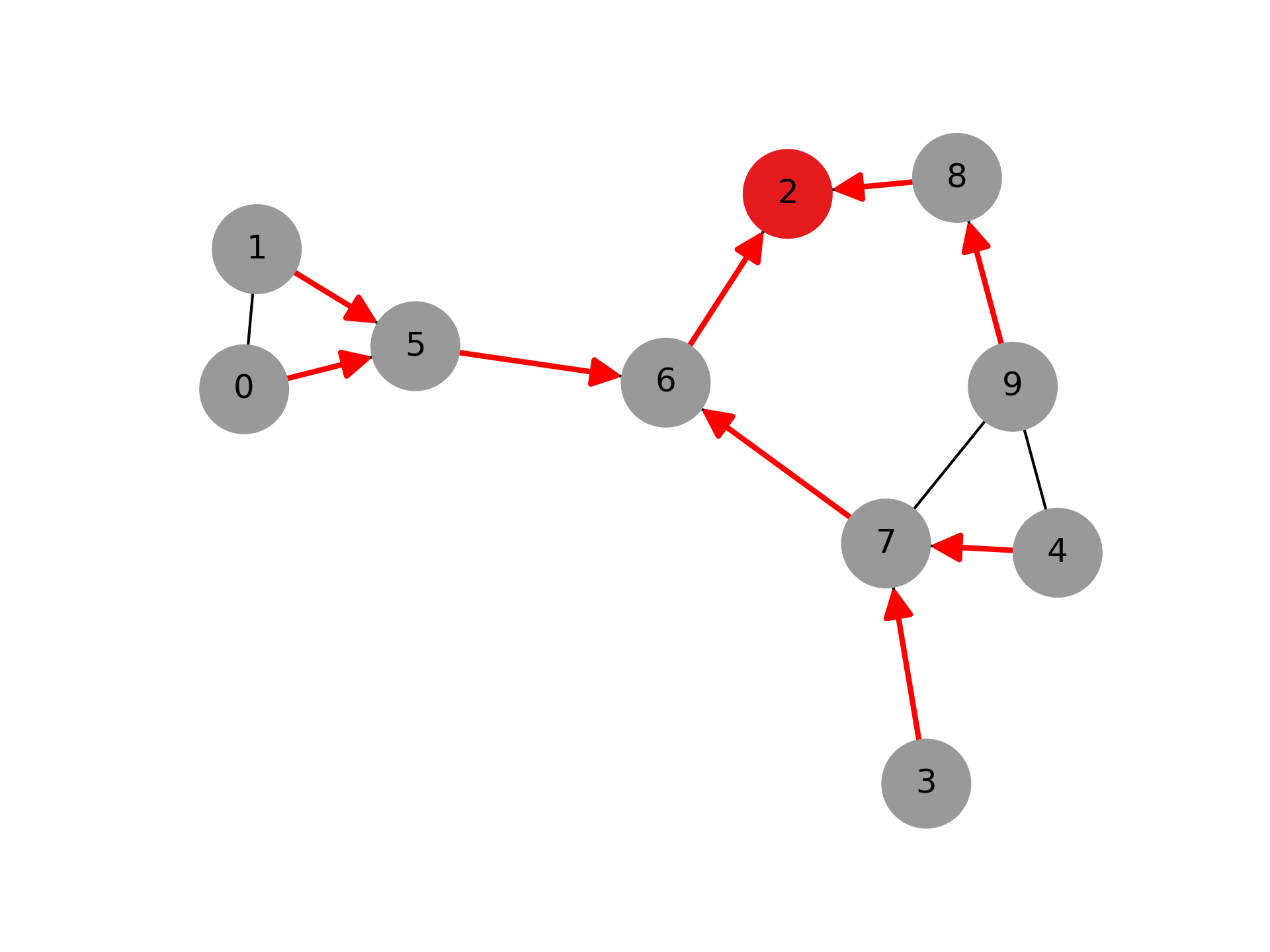}
    \caption{Example of BFS output. The red node corresponds to the source node, and the red edges to the directed BFS tree.}
    \label{fig:example_bfs}
\end{figure}

\begin{figure}
    \centering
    \includegraphics[width=0.8\columnwidth]{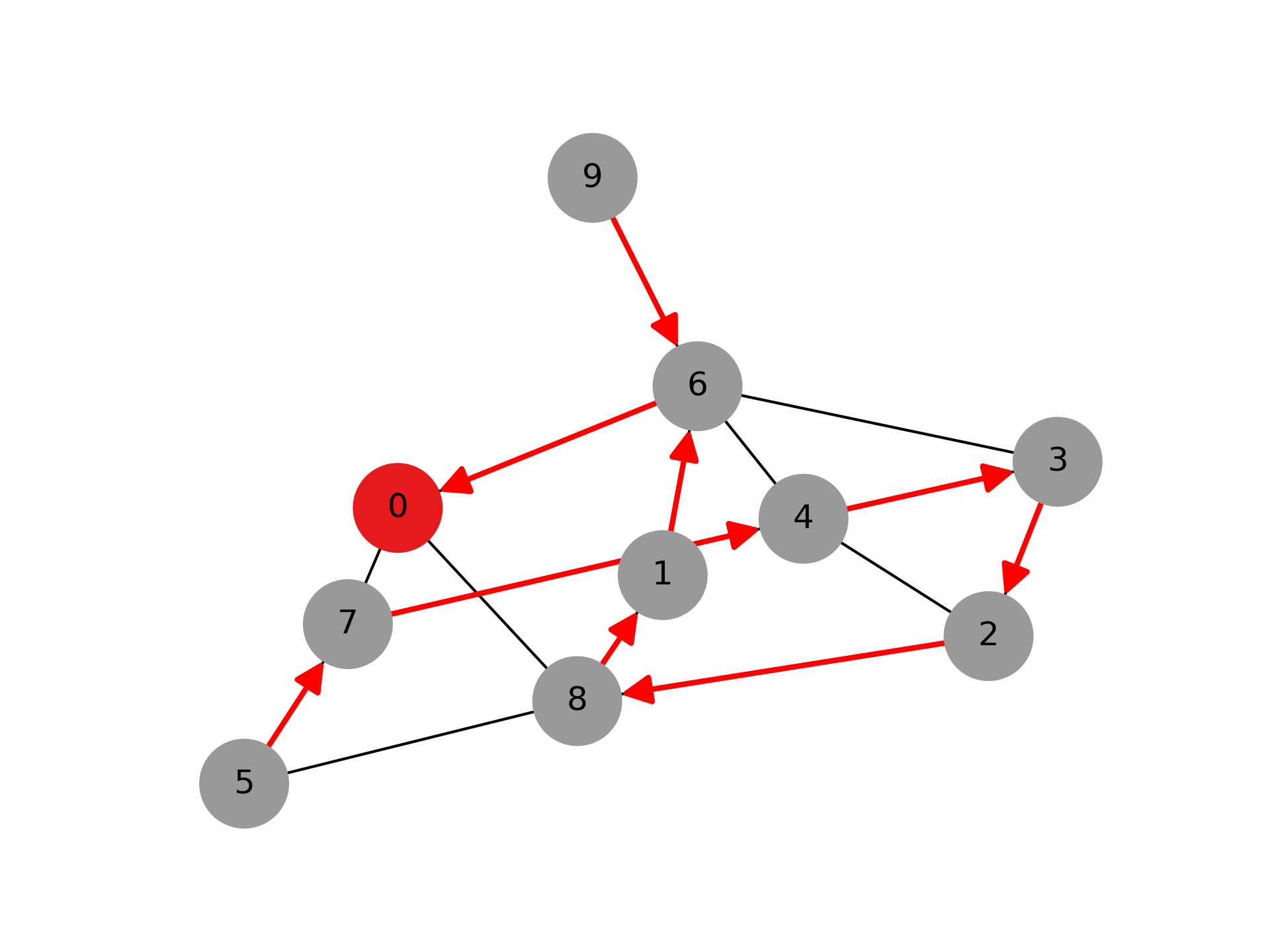}
    \caption{Example of DFS output. The red node 0 corresponds to the source node, and the red edges to the directed DFS tree.}
    \label{fig:example_dfs}
\end{figure}

\begin{figure}
    \centering
    \includegraphics[width=0.8\columnwidth]{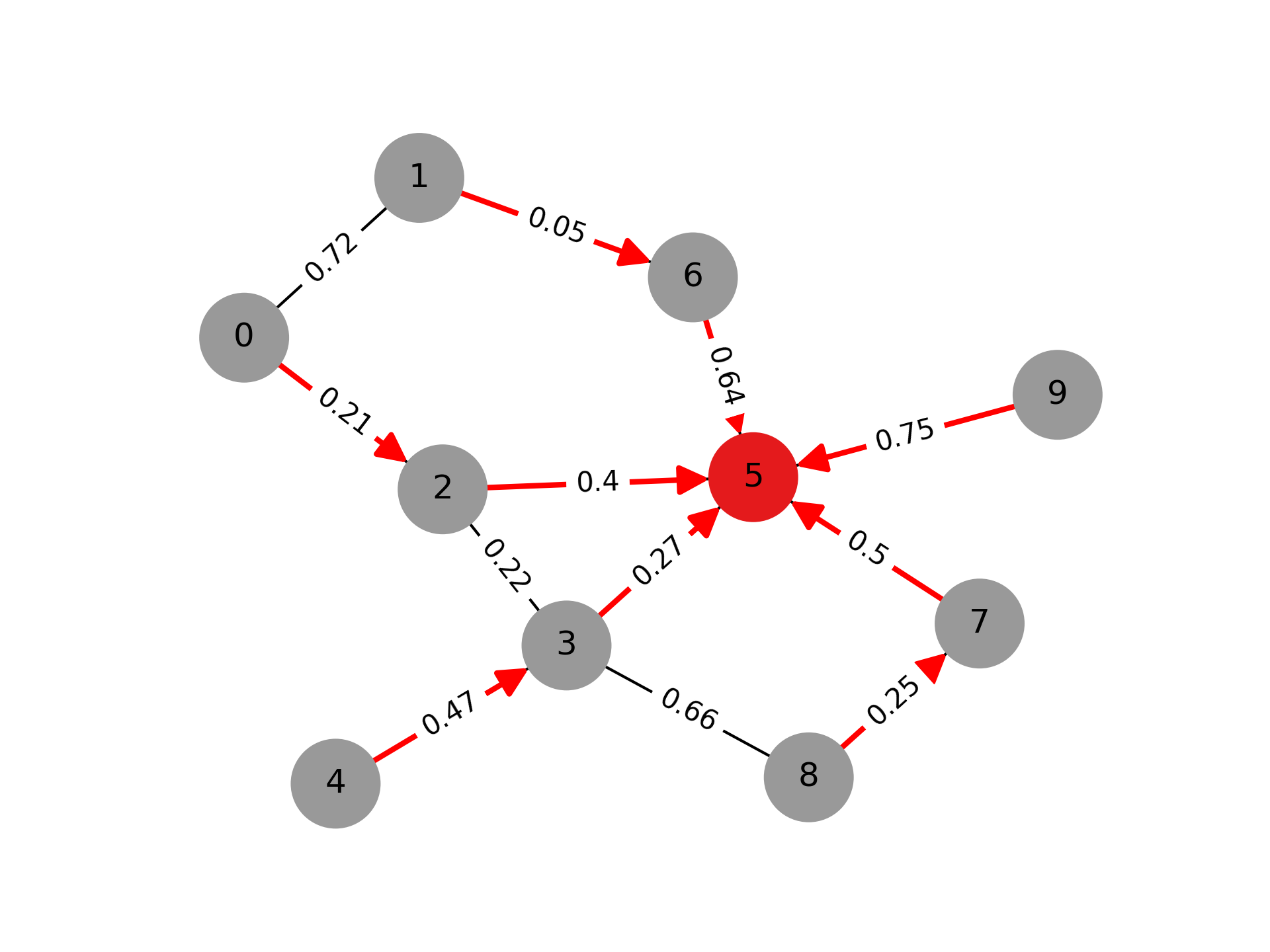}
    \caption{Example of Dijkstra output. The red node corresponds to the source node, and the red edges to the directed shortest path tree.}
    \label{fig:example_dijkstra}
\end{figure}

\begin{figure}
    \centering
    \includegraphics[width=0.8\columnwidth]{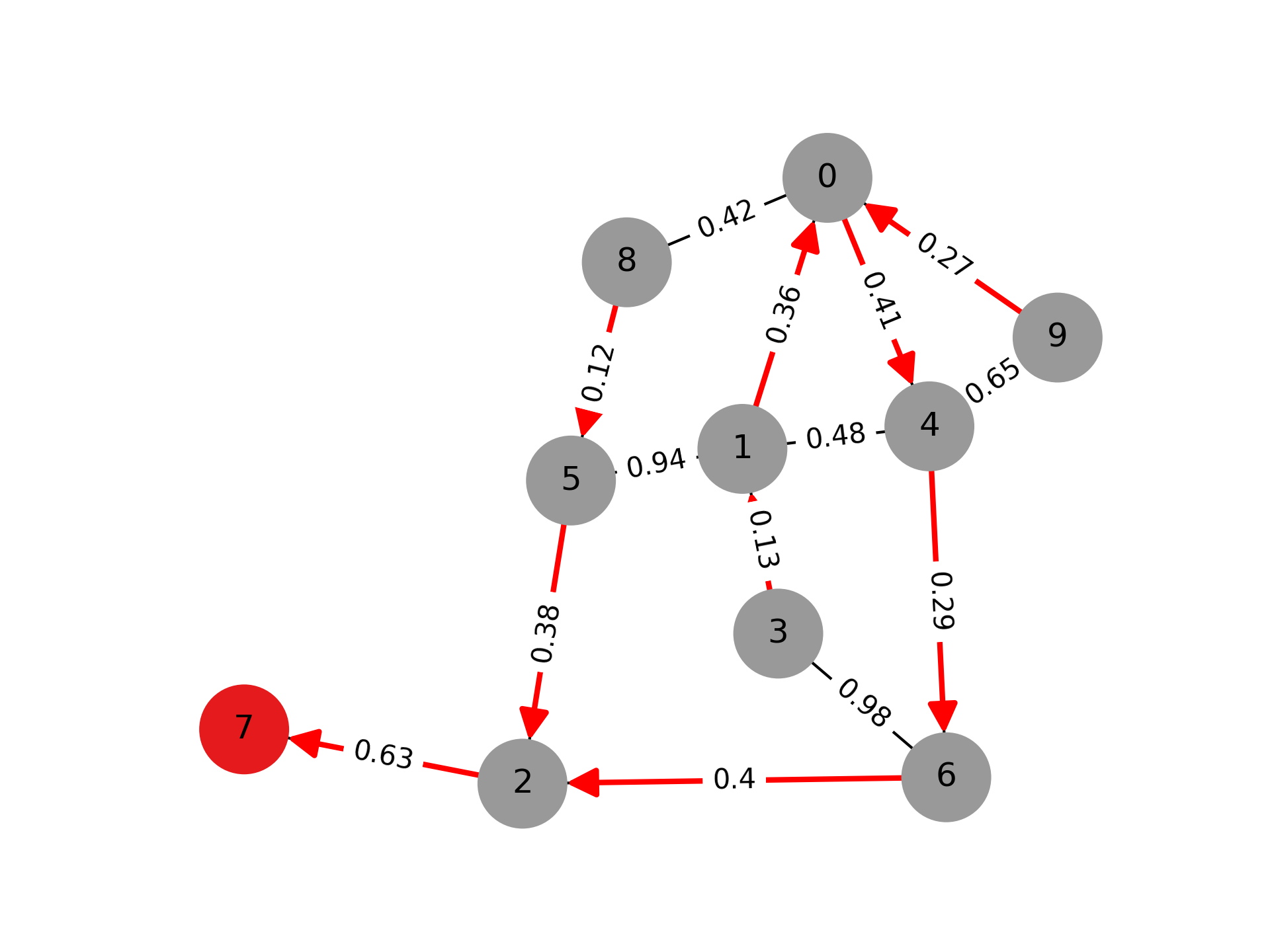}
    \caption{Example of MST output. The red node corresponds to the source node for Prim's algorithm, and the red edges to directed MST.}
    \label{fig:example_mst}
\end{figure}

\begin{figure}
    \centering
    \includegraphics[width=0.8\columnwidth]{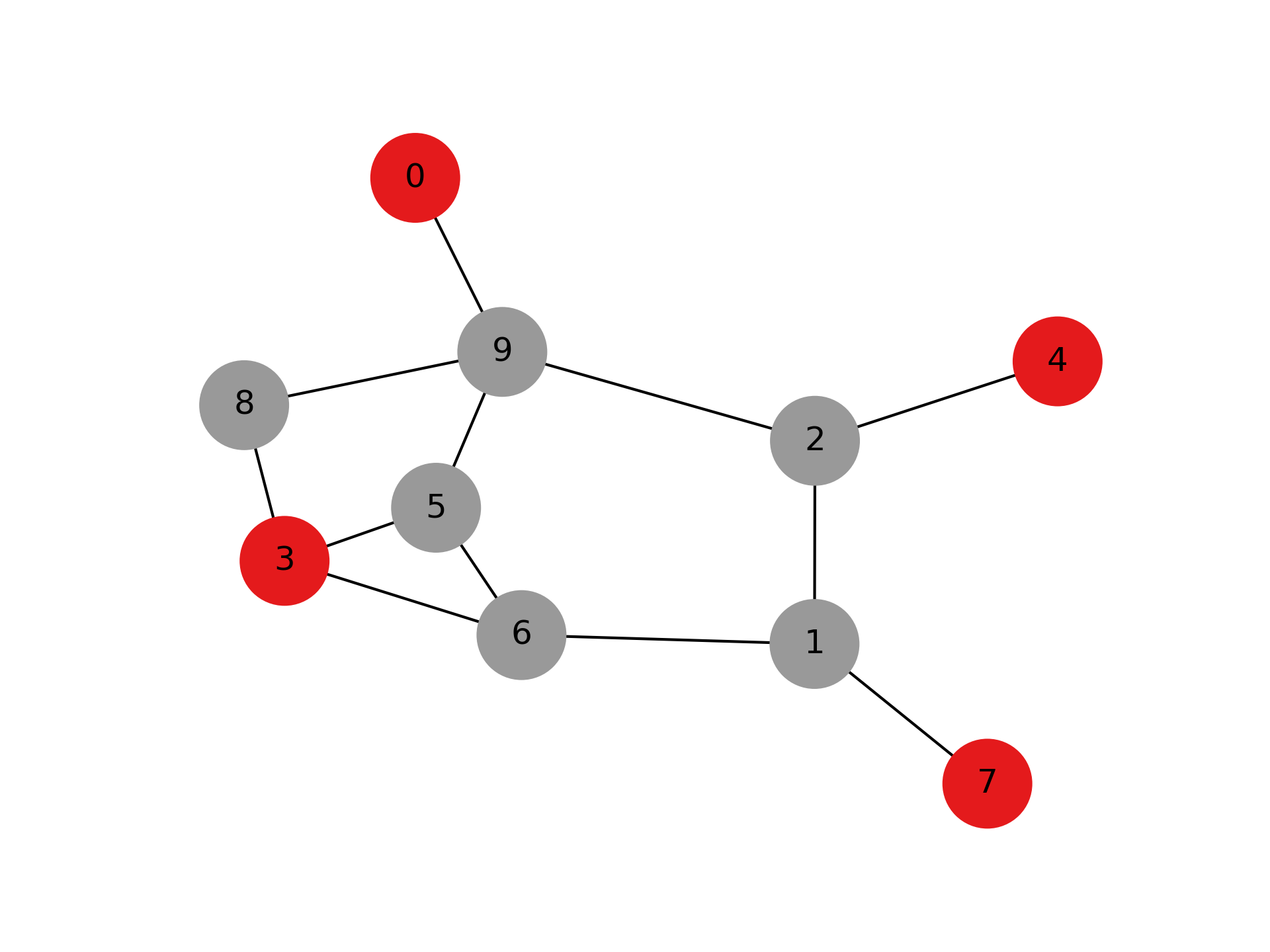}
    \caption{Example of MIS output. The red nodes correspond to the found Maximal Independent Set.}
    \label{fig:example_mis}
\end{figure}

\paragraph{Breadth-first search (BFS)} The input is a pointer to the starting node. The output is the directed BFS tree pointing from node to parent. Refer to Figure \ref{fig:example_bfs} for an example.
\paragraph{Depth-first search (DFS)} The search starts at node 0, and the output is again the directed DFS tree, pointing from leaf to root. Refer to Figure \ref{fig:example_dfs} for an example.
\paragraph{Dijkstra} The Dijkstra shortest path algorithm on a weighted graph. As input, the source node is given. The output is the directed tree that corresponds to the shortest path from all nodes to the source node, again pointing from the leaf to the source node. Refer to Figure \ref{fig:example_dijkstra} for an example.
\paragraph{Maximum Spanning Tree (MST)} Prim's algorithm for finding the Maximum Spanning Tree (MST) of a weighted graph. As input, we are given a source node, and the output is the directed MST pointing from leaf to root. Refer to Figure \ref{fig:example_mst} for an example.
\paragraph{Distributed Maximal Independent Set (MIS)} A Maximal Independent Set within a graph refers to a maximal set of nodes where no two nodes are adjacent. Our implementation is derived from the \emph{Fast MIS} algorithm \cite{wattenhofer2020mastering}. The algorithm relies on \emph{randomness}, enabling a $\mathcal{O}(\log n)$ distributed runtime. The randomness is supplied as an input. The output is a mask over the nodes representing the MIS. Refer to Figure \ref{fig:example_mis} for a visualized example and to Algorithm \ref{alg:mis} for an algorithm outline.

\begin{figure}
    \centering
    \includegraphics[width=0.8\columnwidth]{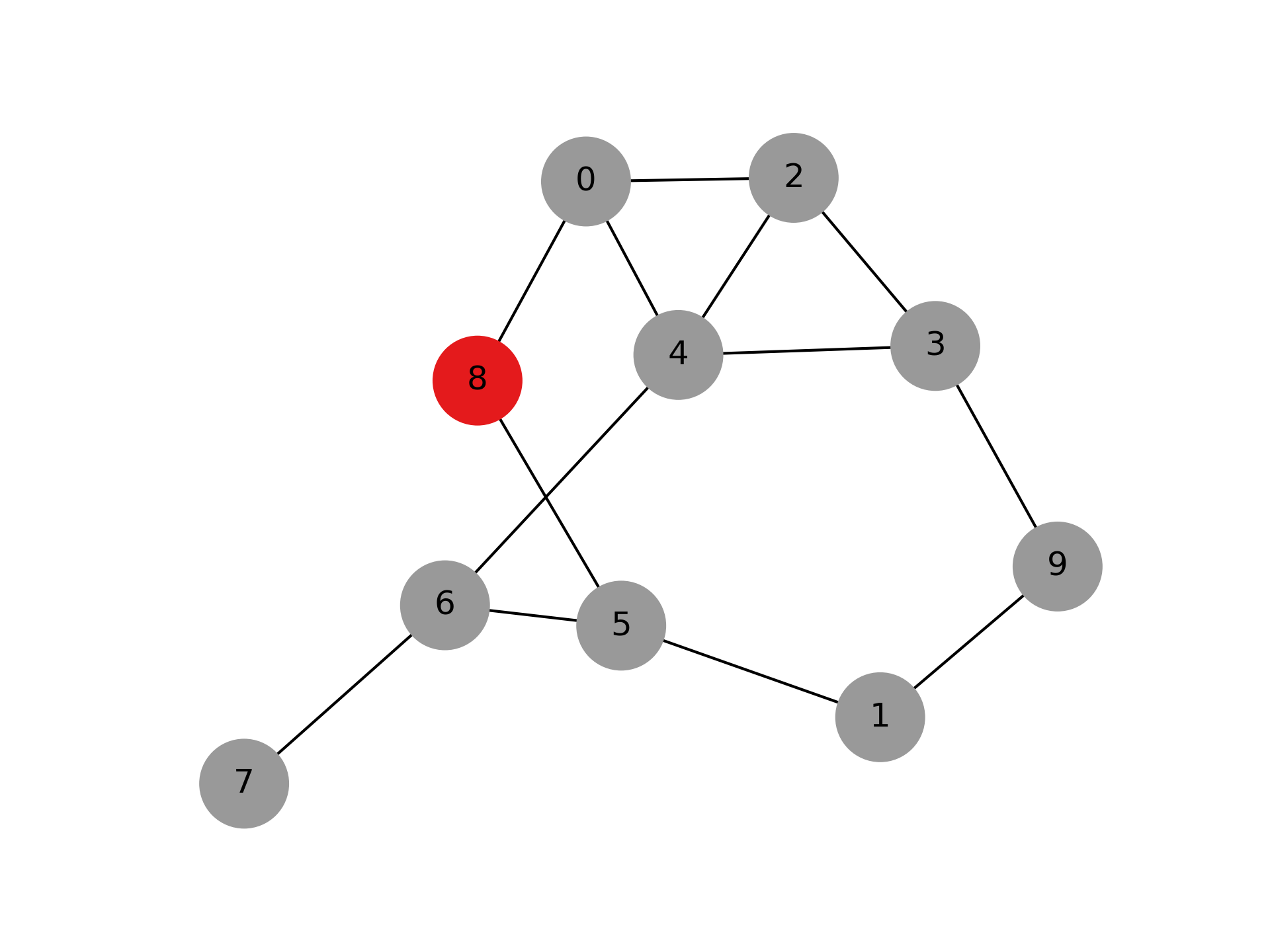}
    \caption{Example of an eccentricity output. The eccentricity of node 8 – the distance to the furthest away node – is 3.} 
    \label{fig:example_eccentricity}
\end{figure}

\paragraph{Distributed Eccentricity} The eccentricity algorithm accepts a graph and a source node as input and produces the source node's eccentricity (also known as radius) as a scalar output. The eccentricity is the maximum distance from the given node to any other node in the graph. It can be solved by a combination of flooding a message through the graph and echoing the maximum value back to the source node. See Algorithm \ref{alg:ecc} for an outline and Figure \ref{fig:example_eccentricity} for a visualized example.

Most of the chosen algorithms for SALSA-CLRS do not explicitly rely on the learned models to perform value generalization by expressing the solution as topological encodings rather than scalar values. However, computing the diameter in the Distributed Eccentricity task requires the models to perform value generalization to larger scalar values for larger graphs. This might be one reason for worse empirical performance compared to other tasks of the SALSA-CLRS benchmark. However, we deem it important for the field of Neural Algorithmic Reasoning overall to consider the challenge of value generalization, which is one of the motivations for expressing the Eccentricity task in this form.

\begin{algorithm}
\caption{Fast MIS 2 \cite{wattenhofer2020mastering}}\label{alg:mis}
\begin{algorithmic}
\State The algorithm operates in synchronous rounds, grouped into phases.
\State \textbf{A single phase} is as follows:
\State \textbf{1)} Each node $\mathit{v}$ takes its precomputed random value $r(v) \in [0,1]$ and sends it to its neighbors.
\State \textbf{2)} If $r(v) < r(w)$ for all neighbors $w$ of $v$, node $v$ enters the MIS and informs its neighbors.
\State \textbf{3)} If $v$ or a neighbor of $v$ entered the MIS, $v$ terminates ($v$ and all edges adjacent to $v$ are removed from the graph), otherwise $v$ enters the next phase.
\end{algorithmic}
\end{algorithm}

\begin{algorithm}
\caption{Eccentricity, adapted from \cite{wattenhofer2020mastering}}\label{alg:ecc}
\begin{algorithmic}
\State The algorithm is a combination of flooding and echoing. Each node can be either dead or alive.
\State \textbf{Flooding:}
\State \textbf{1)} The source node sends the initial flooding message 1 to all neighbors and marks itself dead.
\State \textbf{2)} Each other node $v$, upon receiving the message the first time, increases the message by 1 and forwards it to all alive neighbors. Node $v$ also remembers its parent, the node it got the flooding message from. Once the messages are sent, it marks itself dead. 
\State \textbf{Echoing:}
\State \textbf{3)} If a node receives a flooding message and all of its neighbors are dead, it echos the message back to all of its dead neighbors and removes itself from the graph.
\State \textbf{4)} If a dead node $v$ receives an echo message, it waits until it got an echo from all of its neighbors besides its parent. Then $v$ echos the maximum of the received echos back to its parent. Finally, it removes itself from the graph.
\State \textbf{5)} Once the source node has received an echo from all its children, the maximum received value is its eccentricity.
\end{algorithmic}
\end{algorithm}

\subsection{Dataset} \label{app:ds}
\subsubsection{Graph Types} \label{app:ds:graphs}
\begin{figure}
    \centering
    \includegraphics[width=0.8\columnwidth]{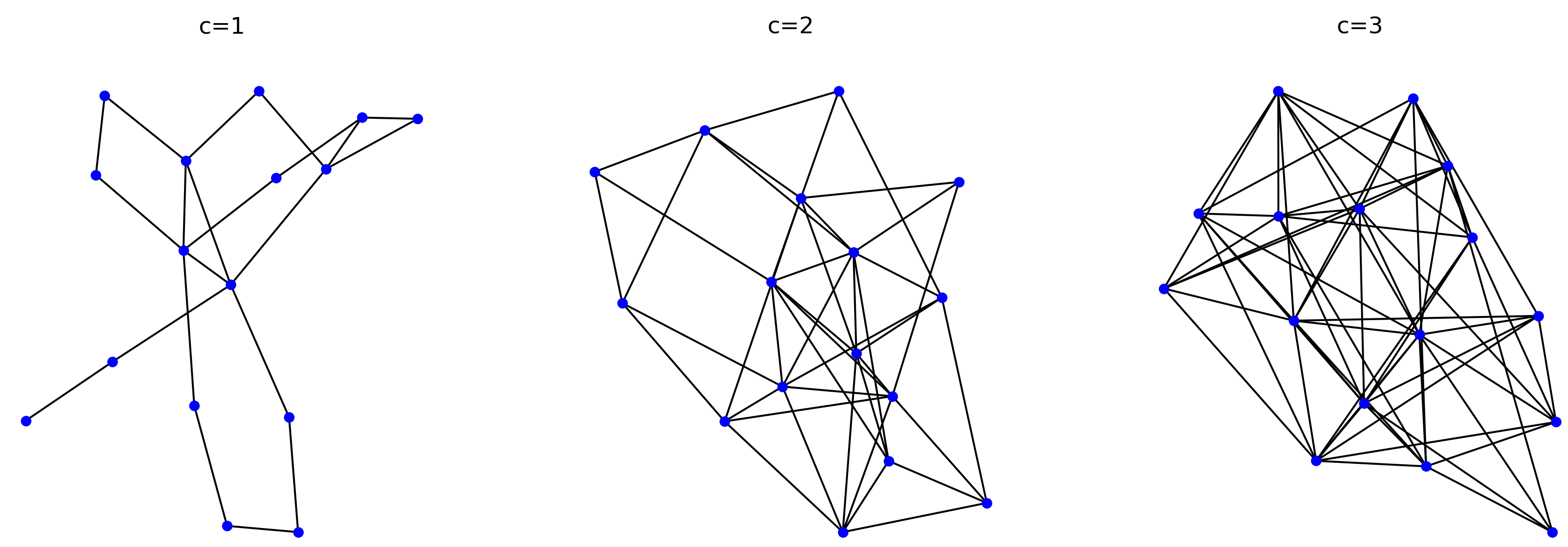}
    \caption{Examples of ER graphs with $n=16$ and $p = c\frac{\ln{n}}{n}\approx 0.173c$.}
    \label{fig:example_er}
\end{figure}

\paragraph{Erdös-Renyi Graphs (ER) \cite{ergraphs}}
An ER graph is generated by choosing each of the  $\frac{n^2-n}{2}$ edges with probability $p$. In CLRS-30 this $p$ is sampled from a range of numbers between $0.01$ and $0.81$\footnote{They sample for the range $[0.1,0.2,...,0.9]$ and square it.}. In ER-graphs the degree of a node grows linearly with the number of nodes in the graph for fixed $p$. This means that when choosing a static $p$, the larger the graph is, the higher its connectivity. Different from CLRS-30, we ensure connectedness on all graphs. As we are interested in sparse graphs, we choose $p$ to be a function of the number of nodes $n$. Hence, we require $p$ to be as low as possible while the graph still remains connected with a high probability. Erdös and Renyi showed that for $c \leq \frac{1}{2}$ for ER graphs $G$ with number of edges $E(G)\approx c\frac{\ln{n}}{n}$ the graph $G$ is \emph{almost surely connected} \cite{erdHos1960evolution}. As the average number of edges in an ER graph is $E(G)\approx { n\choose 2 }p$ we choose $p$ to be \[
p=c\frac{\ln{n}}{n}\] where c is a scalar that is randomly sampled out of the interval $(1,2)$ to increase the diversity in dataset. See Figure \ref{fig:example_er} for examples.

\begin{figure}
    \centering
    \includegraphics[width=0.8\columnwidth]{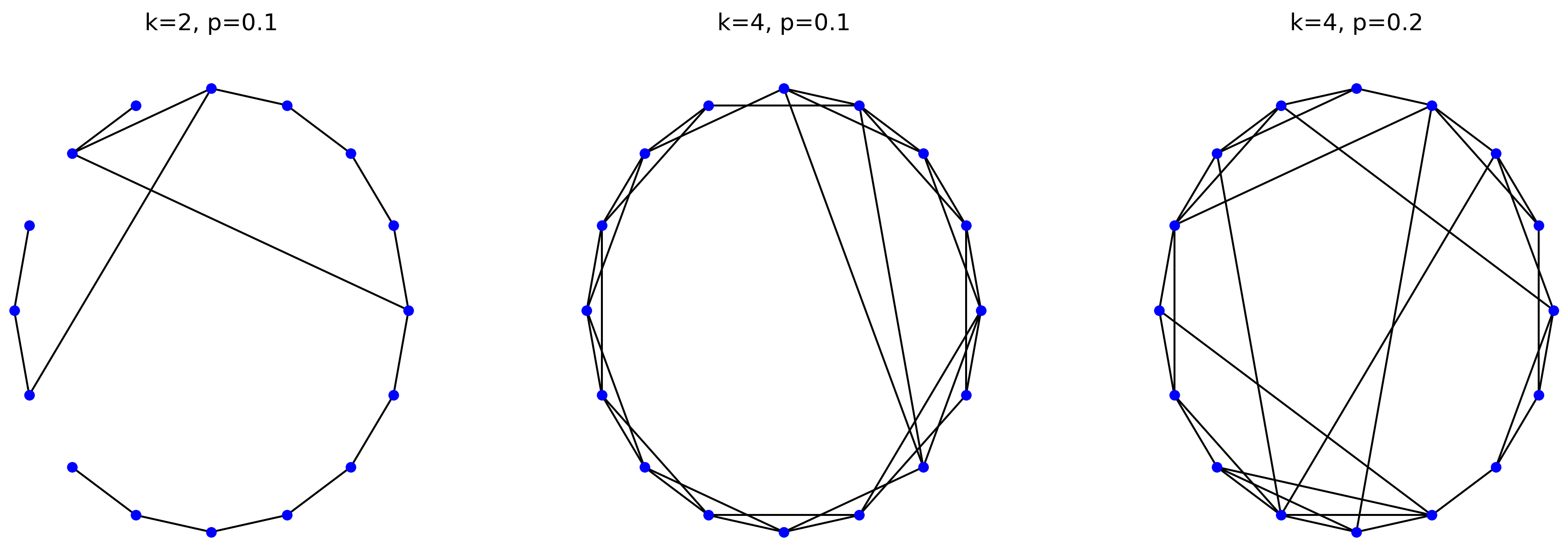}
    \caption{Examples of WS graphs with $n=16$.}
    \label{fig:example_ws}
\end{figure}

\paragraph{Watt Strogatz Graphs (WS) \cite{wsgraphs}} A WS graph, a Small World graph \cite{barthelemy1999small}, is created by taking a ring lattice – a ring where each node is connected to $k$ neighbors ($\frac{k}{2}$ neighbors on each side) – and rewire each edge with probability $p$ to a random other node.  WS graphs have the characteristic that they are globally connected but show local clustering. We also enforce connectivity on WS graphs. As $p\rightarrow \infty$ WS graphs approach ER graphs with $p\approx { n\choose 2 }^{-1}2nk$, so we keep $p$ relatively small. We randomly sample $k$ from $[4,6,8]$ and $p$ from the interval of $(0.05, 0.2)$. See Figure \ref{fig:example_ws} for examples.

\begin{figure}
    \centering
    \includegraphics[width=0.8\columnwidth]{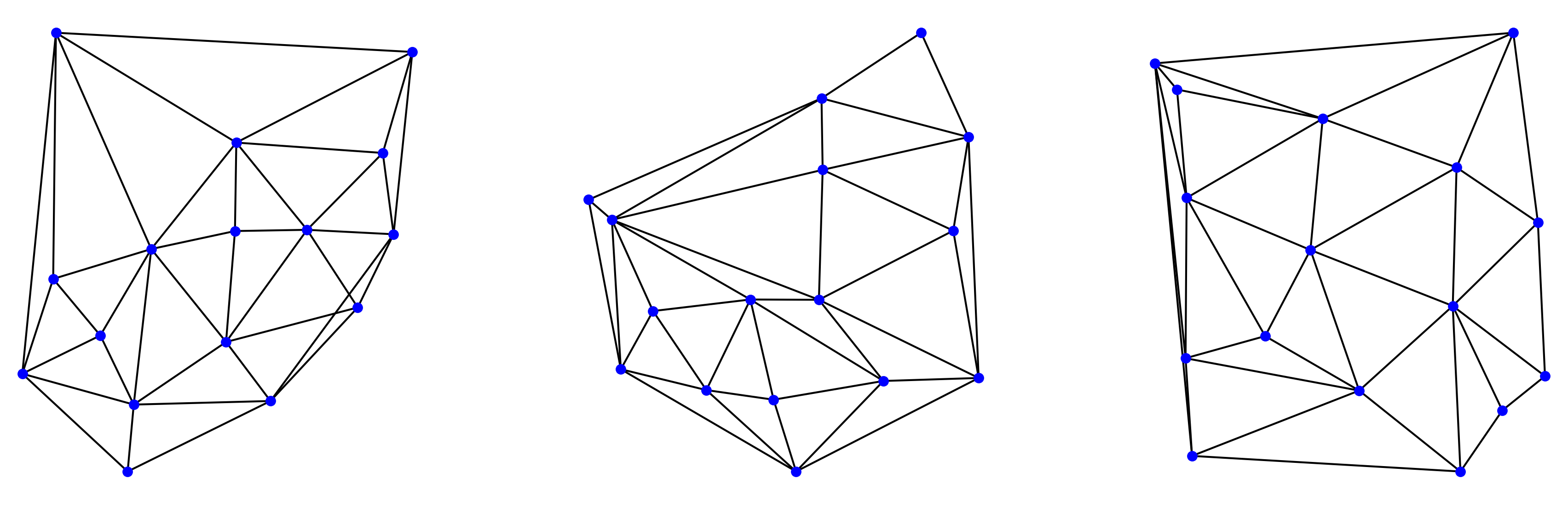}
    \caption{Examples of Delaunay graphs with $n=16$.}
    \label{fig:example_delaunay}
\end{figure}

\paragraph{Delaunay Graphs} Delaunay graphs are created by sampling $n$ points in the plane and computing the Delaunay triangulation. As the graph of a Delaunay triangulation is planar, its average degree is below $6$. See Figure \ref{fig:example_delaunay} for examples.

\subsubsection{Dataset Statistics} \label{app:ds:stats}

 SALSA-CLRS provides a training set, a validation set, and 15 different test sets. The training and validation sets follow closely the datasets CLRS-30, ER graphs with $n$ sampled randomly from $[4, 7, 11, 13, 16]$. We slightly modify this and choose $p$ as described in Section \ref{app:ds:graphs} and increase the size of the training and validation sets to 10000 and 1000, respectively.

The SALSA-CLRS dataset test extrapolation for graphs of size $5\times, 10\times, 50 \times$ and $100\times$ the size of the training graphs, resulting in graphs of size 16, 80, 160, 800, and 1600. This results in 15 different test sets across the three different graph types to measure different facets of OOD performance.

\subsection{Evaluation} \label{app:eval}
\subsubsection{Architectures} \label{app:eval:arch}
We use the same Encode-Process-Decode framework \cite{battaglia2018relational} that CLRS builds on. An encoder – specific to every input feature – produces a latent representation of the input features. These latent representations are aggregated (max) to yield a 128-dimensional node hidden state. Similarly, a per-feature decoder computes the features from the node's hidden state. Notably, when dealing with hints, our model diverges from the CLRS approach. While CLRS decodes the predicted hints in each algorithmic step, calculates a loss, and then re-encodes these hints, our approach streamlines this process. We only decode the hint predictions to calculate the loss and do not re-encode the decoded hints afterward. In the process step, a message-passing layer updates the node embeddings.  As part of the process step, node hidden states undergo an update via a message-passing layer. In scenarios necessitating randomness, the precomputed randomness is concatenated to the processor input. Hints and outputs are decoded from the last two hidden states as well as the input state.

The following three processor modules are evaluated, all  employing maximum aggregation and layer normalization \cite{ba2016layer}. We define $h_v^t$ to be the hidden state of node $v$ at timestep $t$ and  $\mathcal{F}$ to be the aggregation function.
\paragraph{GIN(E)} Standard GIN module with a two-layer Multi-Layer Perceptron (MLP) with ReLu activations and batch norm. We also add a GRU (Gated Recurrent Unit) Cell \cite{cho2014gru} after the message passing to improve training stability. The update without edge weights is defined as:
\[
 h_v^{t+1} = \text{GRU}\left[\Theta_1 \left( (1+\epsilon) \cdot h_v^t + \underset{w \in N(v)}{\mathcal{F}}  h_w^t\right) , h_v^{t} \right]
\]

\paragraph{RecGNN} The architecture proposed by \cite{RecNN}, originally named RecGRU-E. Before the message passing step, an MLP is applied on the edges – the concated node embedding. After each message passing update all node embeddings are passed through a GRU cell. For algorithms that requsire edge weights, we concat the edge weight to each message before we pass it through the edge MLP $\Theta$. The update without edge weights is defined as:
\[
 h_v^{t+1} = \text{GRU}\left[\left(\underset{w \in N(v)}{\mathcal{F}}\Theta\left(h_v^t\middle\| h_w^t\right)\right), h_v^t \right]
\]
\paragraph{PGN} The PGN architecture is introduced in \cite{velivckovic2020pointer}. It defines the following components: The source node linear layer $\Theta_{s}$, the target node linear layer $\Theta_{t}$, the two layer message MLP $\Theta_{msg}$, the skip connection linear layer $\Theta_{skip}$, the output linear layer $\Theta_{out}$ and a ReLu activation $\sigma$. The update without edge weights is defined as

\[
 h_v^{t+1} = \sigma \left(\Theta_{skip}(h_v^{t}) +\Theta_{out}\left[\underset{w \in N(v)}{\mathcal{F}} \Theta_{msg}\left(\Theta_{s}(h_v^{t}) + \Theta_{t}(h_w^{t})\right)\right]\right)
\]

\subsubsection{Experiments} \label{app:eval:exp}
We use early stopping with patience 30. The patience is kept this high because we observed that some training runs dip quite strongly before finding a new optimum. Further, we use a plateau scheduler with patience 10 and factor 0.1. The seeds selected are $42-46$. To combat exploding gradients we apply gradient clipping on the 2-norm of the weights. Adding 2-norm regularization on the hidden node states also helps with training stability. We employ different learning rates for the baseline models, determined by a hyperparameter sweep on the Dijkstra algorithm. For GIN(E) and PGN, we use a learning rate of $0.0004239$, and for RecGNN we use $0.0008$. 

\subsubsection{Metrics}
For all problems, we report Graph Accuracy, referring to whether a graph was entirely solved correctly or not. On the problems BFS, DFS, MIS, Dijkstra, and MST we report Node Accuracy, and for MIS we also report Node F1. As for Eccentricity, we predict a single scalar for the whole graph, the notion of Node Accuracy and Node F1 is not applicable. To compute the Graph Accuracy for Eccentricity, we round the predicted scalar and check whether it was correctly predicted. Additionally, we report the Mean-Squared-Error (MSE) of the unrounded scalar prediction. For all problems predicting a tree with Node Pointers\footnote{A Node Pointer is a data type that serves as a reference to a single neighbor from among the various neighbors connected to a node. To illustrate this concept, consider the case of a BFS tree (the result of the BFS algorithm) in a graph, where a Node Pointer can be used to denote the edge leading to the parent node. We refer to the original CLRS paper for more details.} (BFS, DFS, Dijkstra, MST) each node either predicts its parent correctly or not. An F1 score is not of interest here. It is important to highlight the reporting of all of these metrics and their differences. The Node metrics can be misleading, as is apparent when comparing, e.g., the Node Accuracy performance to the Graph Accuracy. If we predict a node mask, like in MIS, the Node F1 score is the most indicative, as the Node Accuracy does not consider class imbalance.

\subsubsection{Scalability}
\begin{figure}
    \centering
    \includegraphics[width=1.0\linewidth]{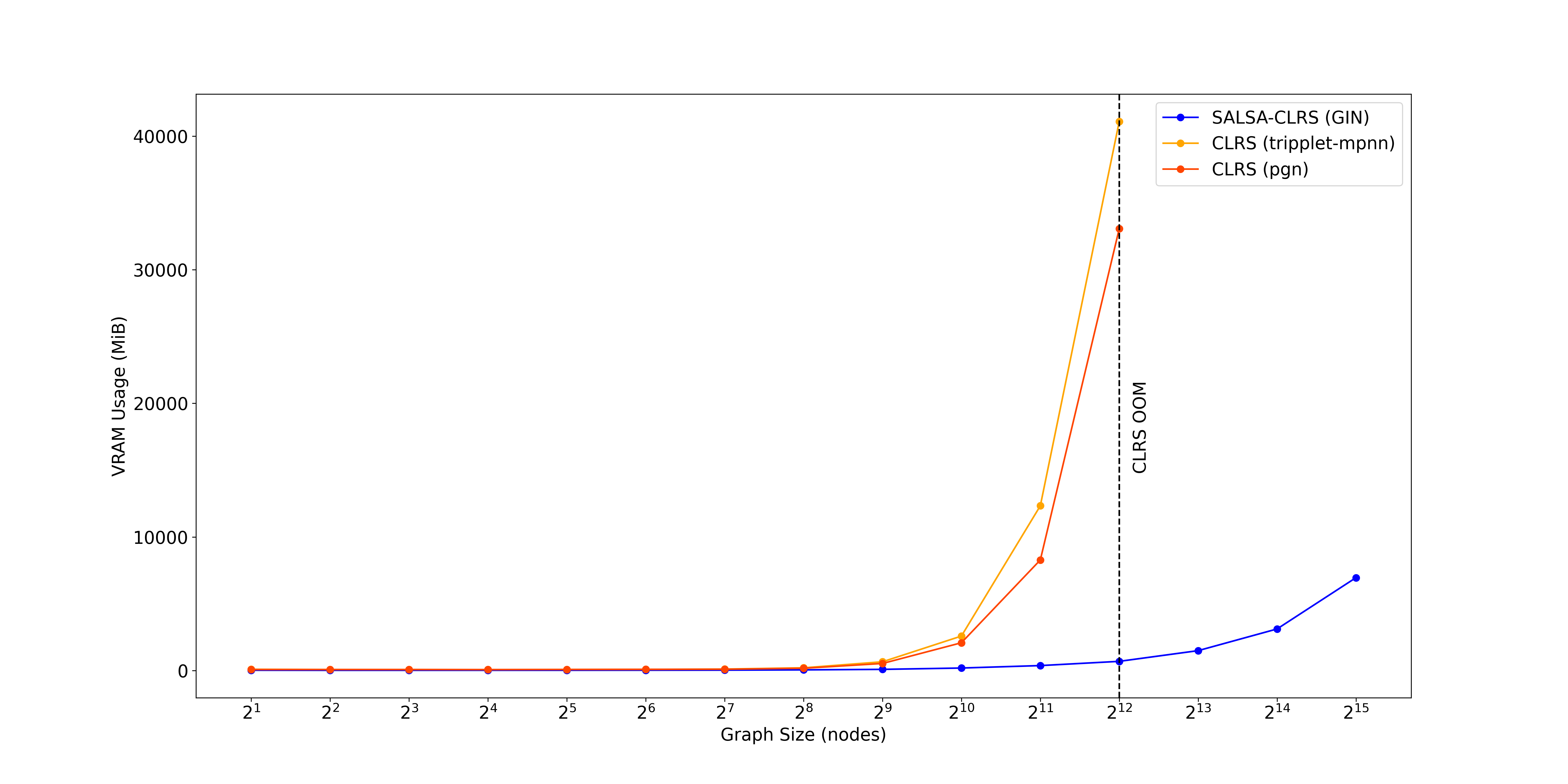}
    \caption{Scalability analysis of SALSA-CLRS and CLRS. Shown is the peak VRAM usage on a NVIDIA A100 GPU with 80GB VRAM across 10 randomly sampled graphs with batch size 1 on the BFS algorithm.}
    \label{fig:vram_usage}
\end{figure}

In Figure \ref{fig:vram_usage}, we assess the scalability of SALSA-CLRS and CLRS by examining their GPU VRAM utilization when used with the BFS algorithm. The figure shows the clear \textbf{asymptotic} advantage brought by the sparsification in SALSA-CLRS. SALSA-CLRS manages inference of graphs as large as 32768 nodes with less than 8GB of VRAM.

To obtain our results, we generated 10 graphs for each graph size using the default settings for each benchmark. For more details regarding these graph types, please refer to Section \ref{subseq:graphtypes}. For SALSA-CLRS, we employed the GIN architecture, while for CLRS, we compared the performance of both the "trippled-mpnn" and "pgn" models. We processed all 10 graphs individually with a batch size of 1. These measurements were conducted on an Nvidia A100 GPU boasting 80GB of VRAM.

Notably, CLRS is implemented in JAX \cite{jax2018github}, whereas SALSA-CLRS is based on PyTorch \cite{Paszke_PyTorch_An_Imperative_2019}, each of which reports memory usage differently. For CLRS, we report memory usage as \code{jax.local\_devices()[gpu\_id].memory\_stats()['peak\_bytes\_in\_use']}, while for SALSA-CLRS, we use \code{torch.cuda.max\_memory\_allocated()}. These reported values might be notably lower than what is indicated by the \emph{nvidia-smi} tool, as the latter includes memory that is reserved but not actively allocated.

It's also worth noting that CLRS uses in-memory datasets, meaning that the entire dataset is stored in RAM. While this approach offers runtime advantages, it can become a bottleneck when dealing with large graphs and datasets due to high RAM usage. For example, a dataset with 1000 graphs of size 1600 does not fit on a machine with 64GB of RAM. In contrast, SALSA-CLRS offloads the dataset to disk and dynamically loads the datapoints into RAM. As long as a single datapoint fits into RAM, the dataset size is not a constraint.

\subsection{Implementation of Node Pointers}
A Node Pointer is used to encode the reference from one node to another and is often used to represent the solutions of the algorithms, i.e., the BFS tree. However, not all fields that are of type \code{(*, Node, Pointer)} behave exactly the same way. In the algorithms we have selected (BFS, DFS, etc.), these fields share one property: The node pointer always points to a neighboring node. Therefore, in order to derive these node pointers, the computational cost is proportional to the amount of edges in the graph. However, for other algorithms that were part of CLRS-30 but are not yet incorporated into SALSA-CLRS (Toposort, MST Kruskal, etc.), these node pointers are no longer restricted in the same way. They could and must, in certain cases, point to arbitrary nodes in the graph (and not just immediate neighbors). For this, all potential edges that could exist in the graph must be considered -- which is again in order of $\mathcal{O}(n^2)$ and clashes with the idea of sparse computation on the original topology.\\
In the example of topological sort, the \code{topo} feature, a node pointer, represents the output of the algorithm. Each node points to the next element in the topological sort, and the last element points to itself. This definition does not guarantee that the pointers are also part of the sparse graph. On the other hand, in DFS, as it is implemented in CLRS, we compute the DFS tree starting at node $0$. The mentioned output feature \code{pi}, also of type node pointer, represents this tree by node pointers, pointing from child to parent. By definition, these nodes are neighboring, so \code{pi} can be encoded on a sparse graph. Single Source Shortest Path uses topsort as a subroutine, so the hint \code{topo\_h} encodes the output of this topsort. Hence, we run into the same problem as described above. Kruskal uses the \code{pi} variable in the union subroutine, which is not sparsely representable. For strongly connected components, the \code{scc\_id} output feature is a node pointer, pointing to the node with the lowest id in each component. This, again, does not require that this edge exists in the sparse graph.


\begin{sidewaystable}[]
\caption{Node Accuracy scores of both models on all algorithms that support Node Accuracy. The table shows also the standard deviation across the 5 runs. Runs marked with \emph{(H)} are trained with hints. All numbers are given as percentages. }
\label{tab:node_acc_all}
\centering
\scriptsize
\addtolength{\tabcolsep}{-0.5em}
\include{results/node_acc}
\centering
\end{sidewaystable}
\begin{sidewaystable}[]

\caption{Graph Accuracy scores of both models on all algorithms. The table shows also the standard deviation across the 5 runs. Runs marked with \emph{(H)} are trained with hints. All numbers are given as percentages.}
\label{tab:graph_acc_all}
\centering
\scriptsize
\addtolength{\tabcolsep}{-0.5em}
\include{results/graph_accuracy}
\centering
\end{sidewaystable}

\begin{table}[h]
\small
    \centering
     \caption{BFS Results. The table shows also the standard deviation across the 5 runs. Runs marked with \emph{(H)} are trained with hints. All numbers are given as percentages.}
    \label{tab:bfs}
    \begin{subtable}[h]{1.0\textwidth}
        \caption{Graph Accuracy}
      \centering
        \include{results/algorithmwise/bfs_graph_accuracy}
    \end{subtable}
    \hfill
    \begin{subtable}[h]{1.0\textwidth}
         \centering
        \caption{Node Accuracy}
        \include{results/algorithmwise/bfs_node_acc}
     \end{subtable}
\end{table}

\begin{table}[h]
\small
     \caption{DFS Results. The table shows also the standard deviation across the 5 runs. Runs marked with \emph{(H)} are trained with hints. All numbers are given as percentages.}
    \label{tab:dfs}
    \centering
    \begin{subtable}[h]{1.0\textwidth}
        \caption{Graph Accuracy}
      \centering
        \include{results/algorithmwise/dfs_graph_accuracy}
    \end{subtable}
    \hfill
    \begin{subtable}[h]{1.0\textwidth}
         \centering
        \caption{Node Accuracy}
        \include{results/algorithmwise/dfs_node_acc}
     \end{subtable}

\end{table}

\begin{table}[h]
\small
     \caption{Dijkstra Results. The table shows also the standard deviation across the 5 runs. Runs marked with \emph{(H)} are trained with hints. All numbers are given as percentages.}
    \label{tab:dijkstra}

    \centering
    \begin{subtable}[h]{1.0\textwidth}
        \caption{Graph Accuracy}
      \centering
        \include{results/algorithmwise/dijkstra_graph_accuracy}
    \end{subtable}
    \hfill
    \begin{subtable}[h]{1.0\textwidth}
         \centering
        \caption{Node Accuracy}
        \include{results/algorithmwise/dijkstra_node_acc}
     \end{subtable}
\end{table}
\begin{table}[h]
\label{tab:mst}
     \caption{MST Results. The table shows also the standard deviation across the 5 runs. Runs marked with \emph{(H)} are trained with hints. All numbers are given as percentages.}
\small
    \centering
    \begin{subtable}[h]{1.0\textwidth}
        \caption{Graph Accuracy}
      \centering
        \include{results/algorithmwise/mst_prim_graph_accuracy}
    \end{subtable}
    \hfill
    \begin{subtable}[h]{1.0\textwidth}
         \centering
        \caption{Node Accuracy}
        \include{results/algorithmwise/mst_prim_node_acc}
     \end{subtable}
    
\end{table}

\begin{table}[h]
     \caption{MIS Results. The table shows also the standard deviation across the 5 runs. Runs marked with \emph{(H)} are trained with hints. All numbers are given as percentages.}
\label{tab:mis}
\small
    \centering
    \begin{subtable}[h]{1.0\textwidth}
        \caption{Graph Accuracy}
      \centering
        \include{results/algorithmwise/fast_mis_graph_accuracy}
    \end{subtable}
    \hfill
    \begin{subtable}[h]{1.0\textwidth}
         \centering
        \caption{Node Accuracy}
        \include{results/algorithmwise/fast_mis_node_acc}
     \end{subtable}
     \begin{subtable}[h]{1.0\textwidth}
      \centering
        \caption{Node F1}
        \include{results/algorithmwise/fast_mis_node_f1}
     \end{subtable}
\end{table}

\begin{table}[h]
\small
     \caption{Eccentricity Results. The table shows also the standard deviation across the 5 runs. Runs marked with \emph{(H)} are trained with hints. Graph Accuracy is given as percentages and Graph MSE is the \emph{Mean-Squared-Error} (lower is better).}
\label{tab:eccentricity}
    \centering
    \begin{subtable}[h]{1.0\textwidth}
        \caption{Graph Accuracy}
      \centering
        \include{results/algorithmwise/eccentricity_graph_accuracy}
    \end{subtable}
    \hfill
    \begin{subtable}[h]{1.0\textwidth}
         \centering
        \caption{Graph MSE}
        \label{fig:bfs:graph_acc}
        \include{results/algorithmwise/eccentricity_mse}
     \end{subtable}
\end{table}

\end{document}

%% file: results/main_results.tex
\resizebox{.95\textwidth}{!}{
\begin{tabular}{l@{\hspace{1em}}l@{\hspace{1em}}S[table-format=3.1]S[table-format=3.1]S[table-format=3.1]S[table-format=3.1]S[table-format=3.1]@{\hspace{1em}}S[table-format=3.1]S[table-format=3.1]S[table-format=3.1]S[table-format=3.1]S[table-format=3.1]@{\hspace{1em}}S[table-format=3.1]S[table-format=3.1]S[table-format=3.1]S[table-format=3.1]S[table-format=3.1]}
\toprule
{\tiny Graph Type}    &        & \multicolumn{5}{l}{ER} & \multicolumn{5}{l}{WS} & \multicolumn{5}{l}{Delaunay} \\
{\tiny n}    &        &   \multicolumn{1}{l}{16}   &  \multicolumn{1}{l}{80}   &  \multicolumn{1}{l}{160}  &  \multicolumn{1}{l}{800}  &  \multicolumn{1}{l}{1600}  &   \multicolumn{1}{l}{16}   &  \multicolumn{1}{l}{80}   &  \multicolumn{1}{l}{160}  &  \multicolumn{1}{l}{800}  &  \multicolumn{1}{l}{1600} &     \multicolumn{1}{l}{16}   &  \multicolumn{1}{l}{80}   &  \multicolumn{1}{l}{160}  &  \multicolumn{1}{l}{800}  &  \multicolumn{1}{l}{1600} \\
\midrule
\midrule
BFS & GIN(E) &  100.0 &   99.6 &   99.3 &   98.0 &  98.0 &   99.9 &  92.9 &   86.7 &  70.4 &  75.3 &    100.0 &   94.3 &  84.6 &  52.7 &  45.9 \\
{\tiny (Node Acc.)}    & PGN &  100.0 &   99.8 &   99.5 &   99.0 &  98.9 &  100.0 &  95.5 &   88.7 &  75.9 &  80.6 &    100.0 &   98.2 &  90.4 &  53.6 &  40.3 \\
    & RecGNN &  100.0 &   99.8 &   99.5 &   99.3 &  99.2 &  100.0 &  97.8 &   94.2 &  82.2 &  82.1 &    100.0 &   98.5 &  92.0 &  67.1 &  55.6 \\
\midrule
DFS & GIN(E) &   49.3 &   30.6 &   19.7 &   18.1 &  16.5 &   29.7 &  15.9 &   16.8 &  22.3 &  20.1 &     46.7 &   28.0 &  25.1 &  23.4 &  23.2 \\
{\tiny (Node Acc.)}    & PGN &   74.2 &   41.2 &   29.9 &   27.8 &  25.8 &   58.8 &  17.9 &   17.7 &  23.6 &  21.3 &     72.7 &   41.7 &  38.2 &  35.8 &  35.4 \\
    & RecGNN &   33.4 &   28.0 &   18.7 &   18.2 &  16.8 &   22.7 &  15.9 &   16.8 &  21.5 &  19.5 &     32.3 &   26.8 &  25.2 &  24.1 &  24.0 \\
\midrule
Dijkstra & GIN(E) &   98.0 &   89.8 &   84.3 &   75.8 &  72.8 &   95.4 &  85.0 &   79.9 &  61.4 &  52.6 &     97.4 &   81.6 &  70.4 &  46.5 &  39.9 \\
{\tiny (Node Acc.)}    & PGN &   99.6 &   98.6 &   97.2 &   94.1 &  92.2 &   98.3 &  97.1 &   95.4 &  81.8 &  72.5 &     99.5 &   97.6 &  92.4 &  62.7 &  51.0 \\
    & RecGNN &   98.5 &   86.8 &   76.0 &   63.7 &  60.6 &   95.8 &  89.2 &   83.9 &  71.4 &  67.3 &     98.0 &   90.4 &  85.0 &  60.2 &  50.0 \\
\midrule
Eccentricity & GIN(E) &   57.3 &   77.1 &   72.3 &   51.3 &  36.7 &   78.0 &  27.6 &    3.6 &   0.0 &   0.0 &     84.8 &    0.0 &   0.0 &   0.0 &   0.0 \\
{\tiny (Graph Acc.)}    & PGN &  100.0 &  100.0 &  100.0 &  100.0 &  64.6 &  100.0 &  93.8 &  100.0 &  25.6 &   5.2 &    100.0 &  100.0 &  76.9 &   0.0 &   0.0 \\
    & RecGNN &   75.8 &   80.5 &   75.0 &   72.7 &  63.0 &   86.7 &  60.8 &   57.4 &  27.6 &  15.2 &     89.9 &   25.2 &   8.3 &   0.0 &   0.0 \\
\midrule
MIS & GIN(E) &   61.2 &   48.2 &   51.7 &   29.5 &  41.1 &   57.5 &  63.5 &   61.7 &  52.7 &  53.2 &     62.7 &   60.5 &  58.1 &  56.5 &  55.2 \\
{\tiny (Node F1)}    & PGN &   99.6 &   99.1 &   98.9 &   96.4 &  97.3 &   99.4 &  99.0 &   97.7 &  90.8 &  86.1 &     99.8 &   99.6 &  99.6 &  99.1 &  98.6 \\
    & RecGNN &   87.7 &   76.5 &   78.5 &   61.8 &  70.6 &   84.0 &  85.6 &   83.9 &  77.5 &  77.9 &     89.3 &   86.5 &  85.5 &  84.3 &  83.4 \\
\midrule
MST & GIN(E) &   92.6 &   79.1 &   77.6 &   74.5 &  72.9 &   89.6 &  75.3 &   74.4 &  73.0 &  72.8 &     92.8 &   77.4 &  75.8 &  74.8 &  74.7 \\
{\tiny (Node Acc.)}    & PGN &   97.3 &   89.1 &   84.6 &   75.7 &  71.9 &   96.8 &  82.5 &   77.6 &  67.4 &  65.1 &     97.4 &   85.2 &  78.5 &  68.7 &  66.8 \\
    & RecGNN &   94.2 &   70.7 &   66.6 &   58.9 &  56.0 &   92.8 &  67.4 &   62.8 &  53.5 &  52.5 &     94.7 &   69.9 &  62.6 &  52.5 &  50.6 \\
\bottomrule
\end{tabular}
}

%% file: results/node_acc.tex
\resizebox{.9\textwidth}{!}{
\begin{tabular}{l@{\hspace{1em}}l@{\hspace{1em}}S[table-format=3.1]@{\hspace{2.5em}}S[table-format=3.1]@{\hspace{2.5em}}S[table-format=3.1]@{\hspace{2.5em}}S[table-format=3.1]@{\hspace{2.5em}}S[table-format=3.1]@{\hspace{4em}}S[table-format=3.1]@{\hspace{2.5em}}S[table-format=3.1]@{\hspace{2.5em}}S[table-format=3.1]@{\hspace{2.5em}}S[table-format=3.1]@{\hspace{2.5em}}S[table-format=3.1]@{\hspace{4em}}S[table-format=3.1]@{\hspace{2.5em}}S[table-format=3.1]@{\hspace{2.5em}}S[table-format=3.1]@{\hspace{2.5em}}S[table-format=3.1]@{\hspace{2.5em}}S[table-format=3.1]@{\hspace{2.5em}}}
\toprule
    &        & \multicolumn{5}{l}{ER} & \multicolumn{5}{l}{WS} & \multicolumn{5}{l}{Delaunay} \\
    &        &   \multicolumn{1}{l}{16}   &  \multicolumn{1}{l}{80}   &  \multicolumn{1}{l}{160}  &  \multicolumn{1}{l}{800}  &  \multicolumn{1}{l}{1600}  &   \multicolumn{1}{l}{16}   &  \multicolumn{1}{l}{80}   &  \multicolumn{1}{l}{160}  &  \multicolumn{1}{l}{800}  &  \multicolumn{1}{l}{1600} &     \multicolumn{1}{l}{16}   &  \multicolumn{1}{l}{80}   &  \multicolumn{1}{l}{160}  &  \multicolumn{1}{l}{800}  &  \multicolumn{1}{l}{1600} \\
\midrule
\midrule
BFS & GIN(E) &  100.0 \text{\tiny $\pm$ 0.1} &   99.6 \text{\tiny $\pm$ 0.4} &   99.3 \text{\tiny $\pm$ 0.6} &   98.0 \text{\tiny $\pm$ 1.6} &   98.0 \text{\tiny $\pm$ 1.5} &   99.9 \text{\tiny $\pm$ 0.3} &   92.9 \text{\tiny $\pm$ 4.2} &   86.7 \text{\tiny $\pm$ 5.5} &  70.4 \text{\tiny $\pm$ 10.8} &   75.3 \text{\tiny $\pm$ 6.1} &  100.0 \text{\tiny $\pm$ 0.1} &   94.3 \text{\tiny $\pm$ 5.6} &  84.6 \text{\tiny $\pm$ 10.6} &  52.7 \text{\tiny $\pm$ 17.2} &  45.9 \text{\tiny $\pm$ 15.8} \\
\bottomrule
\end{tabular}
}

%% file: results/graph_accuracy.tex
\resizebox{.9\textwidth}{!}{
\begin{tabular}{l@{\hspace{1em}}l@{\hspace{1em}}S[table-format=3.1]@{\hspace{2.5em}}S[table-format=3.1]@{\hspace{2.5em}}S[table-format=3.1]@{\hspace{2.5em}}S[table-format=3.1]@{\hspace{2.5em}}S[table-format=3.1]@{\hspace{4em}}S[table-format=3.1]@{\hspace{2.5em}}S[table-format=3.1]@{\hspace{2.5em}}S[table-format=3.1]@{\hspace{2.5em}}S[table-format=3.1]@{\hspace{2.5em}}S[table-format=3.1]@{\hspace{4em}}S[table-format=3.1]@{\hspace{2.5em}}S[table-format=3.1]@{\hspace{2.5em}}S[table-format=3.1]@{\hspace{2.5em}}S[table-format=3.1]@{\hspace{2.5em}}S[table-format=3.1]@{\hspace{2.5em}}}
\toprule
    &        & \multicolumn{5}{l}{ER} & \multicolumn{5}{l}{WS} & \multicolumn{5}{l}{Delaunay} \\
    &        &   \multicolumn{1}{l}{16}   &  \multicolumn{1}{l}{80}   &  \multicolumn{1}{l}{160}  &  \multicolumn{1}{l}{800}  &  \multicolumn{1}{l}{1600}  &   \multicolumn{1}{l}{16}   &  \multicolumn{1}{l}{80}   &  \multicolumn{1}{l}{160}  &  \multicolumn{1}{l}{800}  &  \multicolumn{1}{l}{1600} &     \multicolumn{1}{l}{16}   &  \multicolumn{1}{l}{80}   &  \multicolumn{1}{l}{160}  &  \multicolumn{1}{l}{800}  &  \multicolumn{1}{l}{1600} \\
\midrule
\midrule
BFS & GIN(E) &   99.4 \text{\tiny $\pm$ 0.8} &  84.3 \text{\tiny $\pm$ 13.9} &  57.5 \text{\tiny $\pm$ 15.3} &    2.2 \text{\tiny $\pm$ 4.1} &    0.1 \text{\tiny $\pm$ 0.2} &   98.0 \text{\tiny $\pm$ 4.2} &    5.7 \text{\tiny $\pm$ 8.7} &    0.2 \text{\tiny $\pm$ 0.5} &    0.0 \text{\tiny $\pm$ 0.0} &    0.0 \text{\tiny $\pm$ 0.0} &   99.3 \text{\tiny $\pm$ 1.0} &  25.1 \text{\tiny $\pm$ 28.6} &    0.7 \text{\tiny $\pm$ 1.4} &   0.0 \text{\tiny $\pm$ 0.0} &   0.0 \text{\tiny $\pm$ 0.0} \\
\bottomrule
\end{tabular}

}

%% file: results/algorithmwise/bfs_graph_accuracy.tex
\begin{tabular}{llllllll}
\toprule
         &      &                           GIN &                       GIN (H) &                           PGN &                       PGN (H) &                        RecGNN &                    RecGNN (H) \\
\midrule
ER & 16   &   99.4 \text{\tiny $\pm$ 0.8} &  92.5 \text{\tiny $\pm$ 13.9} &  100.0 \text{\tiny $\pm$ 0.0} &  100.0 \text{\tiny $\pm$ 0.0} &   99.9 \text{\tiny $\pm$ 0.2} &   99.9 \text{\tiny $\pm$ 0.1} \\
         & 80   &  84.3 \text{\tiny $\pm$ 13.9} &  59.4 \text{\tiny $\pm$ 38.3} &   88.7 \text{\tiny $\pm$ 5.9} &   88.1 \text{\tiny $\pm$ 3.8} &   87.9 \text{\tiny $\pm$ 8.8} &  81.7 \text{\tiny $\pm$ 13.0} \\
         & 160  &  57.5 \text{\tiny $\pm$ 15.3} &  37.8 \text{\tiny $\pm$ 37.9} &  54.9 \text{\tiny $\pm$ 21.5} &   66.3 \text{\tiny $\pm$ 8.7} &  55.8 \text{\tiny $\pm$ 24.8} &  49.6 \text{\tiny $\pm$ 25.2} \\
         & 800  &    2.2 \text{\tiny $\pm$ 4.1} &    0.9 \text{\tiny $\pm$ 1.4} &    0.2 \text{\tiny $\pm$ 0.1} &    0.2 \text{\tiny $\pm$ 0.3} &    4.6 \text{\tiny $\pm$ 6.5} &    1.8 \text{\tiny $\pm$ 2.3} \\
         & 1600 &    0.1 \text{\tiny $\pm$ 0.2} &    0.0 \text{\tiny $\pm$ 0.1} &    0.0 \text{\tiny $\pm$ 0.0} &    0.0 \text{\tiny $\pm$ 0.0} &    0.4 \text{\tiny $\pm$ 0.6} &    0.0 \text{\tiny $\pm$ 0.1} \\
\midrule
WS & 16   &   98.0 \text{\tiny $\pm$ 4.2} &  92.8 \text{\tiny $\pm$ 12.0} &  100.0 \text{\tiny $\pm$ 0.0} &  100.0 \text{\tiny $\pm$ 0.0} &  100.0 \text{\tiny $\pm$ 0.0} &   99.4 \text{\tiny $\pm$ 1.3} \\
         & 80   &    5.7 \text{\tiny $\pm$ 8.7} &  10.2 \text{\tiny $\pm$ 13.8} &   13.1 \text{\tiny $\pm$ 3.3} &   14.2 \text{\tiny $\pm$ 3.6} &  32.5 \text{\tiny $\pm$ 18.3} &  20.7 \text{\tiny $\pm$ 13.5} \\
\bottomrule
\end{tabular}

%% file: results/algorithmwise/bfs_node_acc.tex
\begin{tabular}{llllllll}
\toprule
         &      &                           GIN &                       GIN (H) &                           PGN &                       PGN (H) &                        RecGNN &                    RecGNN (H) \\
\midrule
ER & 16   &  100.0 \text{\tiny $\pm$ 0.1} &   98.8 \text{\tiny $\pm$ 2.4} &  100.0 \text{\tiny $\pm$ 0.0} &  100.0 \text{\tiny $\pm$ 0.0} &  100.0 \text{\tiny $\pm$ 0.0} &  100.0 \text{\tiny $\pm$ 0.0} \\
         & 80   &   99.6 \text{\tiny $\pm$ 0.4} &   95.3 \text{\tiny $\pm$ 9.2} &   99.8 \text{\tiny $\pm$ 0.1} &   99.8 \text{\tiny $\pm$ 0.1} &   99.8 \text{\tiny $\pm$ 0.1} &   99.6 \text{\tiny $\pm$ 0.2} \\
         & 160  &   99.3 \text{\tiny $\pm$ 0.6} &   95.1 \text{\tiny $\pm$ 8.9} &   99.5 \text{\tiny $\pm$ 0.3} &   99.6 \text{\tiny $\pm$ 0.1} &   99.5 \text{\tiny $\pm$ 0.3} &   99.3 \text{\tiny $\pm$ 0.5} \\
         & 800  &   98.0 \text{\tiny $\pm$ 1.6} &  86.9 \text{\tiny $\pm$ 26.1} &   99.0 \text{\tiny $\pm$ 0.2} &   98.7 \text{\tiny $\pm$ 0.3} &   99.3 \text{\tiny $\pm$ 0.4} &   99.0 \text{\tiny $\pm$ 0.5} \\
         & 1600 &   98.0 \text{\tiny $\pm$ 1.5} &  86.5 \text{\tiny $\pm$ 27.2} &   98.9 \text{\tiny $\pm$ 0.2} &   98.5 \text{\tiny $\pm$ 0.3} &   99.2 \text{\tiny $\pm$ 0.4} &   98.6 \text{\tiny $\pm$ 0.6} \\
\midrule
WS & 16   &   99.9 \text{\tiny $\pm$ 0.3} &   99.2 \text{\tiny $\pm$ 1.4} &  100.0 \text{\tiny $\pm$ 0.0} &  100.0 \text{\tiny $\pm$ 0.0} &  100.0 \text{\tiny $\pm$ 0.0} &  100.0 \text{\tiny $\pm$ 0.1} \\
         & 80   &   92.9 \text{\tiny $\pm$ 4.2} &  83.0 \text{\tiny $\pm$ 25.0} &   95.5 \text{\tiny $\pm$ 0.7} &   96.1 \text{\tiny $\pm$ 0.5} &   97.8 \text{\tiny $\pm$ 1.1} &   96.7 \text{\tiny $\pm$ 0.8} \\
\bottomrule
\end{tabular}

%% file: results/algorithmwise/dfs_graph_accuracy.tex
\begin{tabular}{llllllll}
\toprule
         &      &                         GIN &                     GIN (H) &                           PGN &                       PGN (H) &                      RecGNN &                   RecGNN (H) \\
\midrule
ER & 16   &  0.1 \text{\tiny $\pm$ 0.1} &  0.0 \text{\tiny $\pm$ 0.0} &  18.4 \text{\tiny $\pm$ 37.7} &  19.9 \text{\tiny $\pm$ 30.7} &  0.0 \text{\tiny $\pm$ 0.0} &   4.5 \text{\tiny $\pm$ 7.8} \\
         & 80   &  0.0 \text{\tiny $\pm$ 0.0} &  0.0 \text{\tiny $\pm$ 0.0} &    0.0 \text{\tiny $\pm$ 0.0} &    0.0 \text{\tiny $\pm$ 0.0} &  0.0 \text{\tiny $\pm$ 0.0} &   0.0 \text{\tiny $\pm$ 0.0} \\
         & 160  &  0.0 \text{\tiny $\pm$ 0.0} &  0.0 \text{\tiny $\pm$ 0.0} &    0.0 \text{\tiny $\pm$ 0.0} &    0.0 \text{\tiny $\pm$ 0.0} &  0.0 \text{\tiny $\pm$ 0.0} &   0.0 \text{\tiny $\pm$ 0.0} \\
         & 800  &  0.0 \text{\tiny $\pm$ 0.0} &  0.0 \text{\tiny $\pm$ 0.0} &    0.0 \text{\tiny $\pm$ 0.0} &    0.0 \text{\tiny $\pm$ 0.0} &  0.0 \text{\tiny $\pm$ 0.0} &   0.0 \text{\tiny $\pm$ 0.0} \\
         & 1600 &  0.0 \text{\tiny $\pm$ 0.0} &  0.0 \text{\tiny $\pm$ 0.0} &    0.0 \text{\tiny $\pm$ 0.0} &    0.0 \text{\tiny $\pm$ 0.0} &  0.0 \text{\tiny $\pm$ 0.0} &   0.0 \text{\tiny $\pm$ 0.0} \\
\midrule
WS & 16   &  0.0 \text{\tiny $\pm$ 0.0} &  0.0 \text{\tiny $\pm$ 0.0} &   9.5 \text{\tiny $\pm$ 21.2} &    3.2 \text{\tiny $\pm$ 7.2} &  0.0 \text{\tiny $\pm$ 0.0} &   0.0 \text{\tiny $\pm$ 0.0} \\
         & 80   &  0.0 \text{\tiny $\pm$ 0.0} &  0.0 \text{\tiny $\pm$ 0.0} &    0.0 \text{\tiny $\pm$ 0.0} &    0.0 \text{\tiny $\pm$ 0.0} &  0.0 \text{\tiny $\pm$ 0.0} &   0.0 \text{\tiny $\pm$ 0.0} \\
\bottomrule
\end{tabular}

%% file: results/algorithmwise/dfs_node_acc.tex
\begin{tabular}{llllllll}
\toprule
         &      &                          GIN &                      GIN (H) &                           PGN &                       PGN (H) &                        RecGNN &                    RecGNN (H) \\
\midrule
ER & 16   &  49.3 \text{\tiny $\pm$ 8.1} &  41.5 \text{\tiny $\pm$ 7.5} &  74.2 \text{\tiny $\pm$ 14.0} &   82.0 \text{\tiny $\pm$ 9.2} &  33.4 \text{\tiny $\pm$ 14.5} &  48.3 \text{\tiny $\pm$ 19.1} \\
         & 80   &  30.6 \text{\tiny $\pm$ 4.0} &  30.4 \text{\tiny $\pm$ 2.3} &   41.2 \text{\tiny $\pm$ 3.8} &   38.4 \text{\tiny $\pm$ 2.7} &   28.0 \text{\tiny $\pm$ 6.5} &   22.8 \text{\tiny $\pm$ 4.7} \\
         & 160  &  19.7 \text{\tiny $\pm$ 3.9} &  20.0 \text{\tiny $\pm$ 3.1} &   29.9 \text{\tiny $\pm$ 2.6} &   26.9 \text{\tiny $\pm$ 2.5} &   18.7 \text{\tiny $\pm$ 4.1} &   13.5 \text{\tiny $\pm$ 4.6} \\
         & 800  &  18.1 \text{\tiny $\pm$ 3.8} &  19.5 \text{\tiny $\pm$ 2.6} &   27.8 \text{\tiny $\pm$ 2.1} &   24.9 \text{\tiny $\pm$ 2.3} &   18.2 \text{\tiny $\pm$ 4.4} &   13.1 \text{\tiny $\pm$ 4.1} \\
         & 1600 &  16.5 \text{\tiny $\pm$ 3.5} &  17.8 \text{\tiny $\pm$ 2.5} &   25.8 \text{\tiny $\pm$ 2.1} &   23.1 \text{\tiny $\pm$ 2.3} &   16.8 \text{\tiny $\pm$ 4.3} &   12.0 \text{\tiny $\pm$ 3.6} \\
\midrule
WS & 16   &  29.7 \text{\tiny $\pm$ 4.9} &  25.0 \text{\tiny $\pm$ 3.7} &  58.8 \text{\tiny $\pm$ 20.8} &  57.6 \text{\tiny $\pm$ 17.6} &   22.7 \text{\tiny $\pm$ 8.2} &  35.3 \text{\tiny $\pm$ 17.7} \\
         & 80   &  15.9 \text{\tiny $\pm$ 0.9} &  15.8 \text{\tiny $\pm$ 0.6} &   17.9 \text{\tiny $\pm$ 1.7} &   17.0 \text{\tiny $\pm$ 1.6} &   15.9 \text{\tiny $\pm$ 1.5} &   13.5 \text{\tiny $\pm$ 2.6} \\
\bottomrule
\end{tabular}

%% file: results/algorithmwise/dijkstra_graph_accuracy.tex
\begin{tabular}{llllllll}
\toprule
         &      &                         GINE &                      GINE (H) &                          PGN &                      PGN (H) &                        RecGNN &                   RecGNN (H) \\
\midrule
ER & 16   &  73.4 \text{\tiny $\pm$ 2.6} &  49.8 \text{\tiny $\pm$ 10.8} &  94.6 \text{\tiny $\pm$ 1.1} &  89.5 \text{\tiny $\pm$ 1.0} &  81.7 \text{\tiny $\pm$ 16.1} &  73.8 \text{\tiny $\pm$ 1.6} \\
         & 80   &   0.2 \text{\tiny $\pm$ 0.2} &    0.0 \text{\tiny $\pm$ 0.0} &  37.8 \text{\tiny $\pm$ 6.9} &   3.3 \text{\tiny $\pm$ 3.7} &    6.8 \text{\tiny $\pm$ 6.1} &   0.0 \text{\tiny $\pm$ 0.0} \\
         & 160  &   0.0 \text{\tiny $\pm$ 0.0} &    0.0 \text{\tiny $\pm$ 0.0} &   5.2 \text{\tiny $\pm$ 1.9} &   0.0 \text{\tiny $\pm$ 0.1} &    0.3 \text{\tiny $\pm$ 0.5} &   0.0 \text{\tiny $\pm$ 0.0} \\
         & 800  &   0.0 \text{\tiny $\pm$ 0.0} &    0.0 \text{\tiny $\pm$ 0.0} &   0.0 \text{\tiny $\pm$ 0.0} &   0.0 \text{\tiny $\pm$ 0.0} &    0.0 \text{\tiny $\pm$ 0.0} &   0.0 \text{\tiny $\pm$ 0.0} \\
         & 1600 &   0.0 \text{\tiny $\pm$ 0.0} &    0.0 \text{\tiny $\pm$ 0.0} &   0.0 \text{\tiny $\pm$ 0.0} &   0.0 \text{\tiny $\pm$ 0.0} &    0.0 \text{\tiny $\pm$ 0.0} &   0.0 \text{\tiny $\pm$ 0.0} \\
\midrule
WS & 16   &  51.6 \text{\tiny $\pm$ 3.0} &   28.7 \text{\tiny $\pm$ 9.9} &  76.4 \text{\tiny $\pm$ 4.0} &  70.8 \text{\tiny $\pm$ 2.4} &  60.4 \text{\tiny $\pm$ 22.7} &  50.9 \text{\tiny $\pm$ 5.6} \\
         & 80   &   0.0 \text{\tiny $\pm$ 0.0} &    0.0 \text{\tiny $\pm$ 0.0} &  17.2 \text{\tiny $\pm$ 2.8} &   0.4 \text{\tiny $\pm$ 0.6} &    8.4 \text{\tiny $\pm$ 7.4} &   0.0 \text{\tiny $\pm$ 0.0} \\
\bottomrule
\end{tabular}

%% file: results/algorithmwise/dijkstra_node_acc.tex
\begin{tabular}{llllllll}
\toprule
         &      &                         GINE &                     GINE (H) &                          PGN &                      PGN (H) &                        RecGNN &                    RecGNN (H) \\
\midrule
ER & 16   &  98.0 \text{\tiny $\pm$ 0.2} &  95.2 \text{\tiny $\pm$ 1.8} &  99.6 \text{\tiny $\pm$ 0.1} &  99.3 \text{\tiny $\pm$ 0.1} &   98.5 \text{\tiny $\pm$ 1.6} &   98.0 \text{\tiny $\pm$ 0.1} \\
         & 80   &  89.8 \text{\tiny $\pm$ 1.1} &  62.4 \text{\tiny $\pm$ 7.0} &  98.6 \text{\tiny $\pm$ 0.3} &  94.2 \text{\tiny $\pm$ 2.5} &  86.8 \text{\tiny $\pm$ 15.4} &  32.9 \text{\tiny $\pm$ 21.6} \\
         & 160  &  84.3 \text{\tiny $\pm$ 1.6} &  53.3 \text{\tiny $\pm$ 6.2} &  97.2 \text{\tiny $\pm$ 0.5} &  92.0 \text{\tiny $\pm$ 2.3} &  76.0 \text{\tiny $\pm$ 22.1} &  25.0 \text{\tiny $\pm$ 17.4} \\
         & 800  &  75.8 \text{\tiny $\pm$ 2.2} &  40.4 \text{\tiny $\pm$ 8.1} &  94.1 \text{\tiny $\pm$ 0.6} &  87.1 \text{\tiny $\pm$ 2.7} &  63.7 \text{\tiny $\pm$ 27.7} &  17.7 \text{\tiny $\pm$ 12.2} \\
         & 1600 &  72.8 \text{\tiny $\pm$ 2.3} &  36.9 \text{\tiny $\pm$ 7.6} &  92.2 \text{\tiny $\pm$ 0.7} &  84.5 \text{\tiny $\pm$ 3.4} &  60.6 \text{\tiny $\pm$ 27.7} &  16.4 \text{\tiny $\pm$ 10.7} \\
\midrule
WS & 16   &  95.4 \text{\tiny $\pm$ 0.7} &  91.2 \text{\tiny $\pm$ 3.5} &  98.3 \text{\tiny $\pm$ 0.4} &  97.8 \text{\tiny $\pm$ 0.2} &   95.8 \text{\tiny $\pm$ 4.2} &   95.5 \text{\tiny $\pm$ 1.0} \\
         & 80   &  85.0 \text{\tiny $\pm$ 1.4} &  55.3 \text{\tiny $\pm$ 9.3} &  97.1 \text{\tiny $\pm$ 0.2} &  85.8 \text{\tiny $\pm$ 6.0} &  89.2 \text{\tiny $\pm$ 14.1} &  36.3 \text{\tiny $\pm$ 16.4} \\
\bottomrule
\end{tabular}

%% file: results/algorithmwise/mst_prim_graph_accuracy.tex
\begin{tabular}{llllllll}
\toprule
         &      &                         GINE &                     GINE (H) &                          PGN &                      PGN (H) &                        RecGNN &                   RecGNN (H) \\
\midrule
ER & 16   &  43.2 \text{\tiny $\pm$ 4.6} &  29.7 \text{\tiny $\pm$ 5.6} &  79.2 \text{\tiny $\pm$ 4.3} &  69.9 \text{\tiny $\pm$ 6.1} &  56.8 \text{\tiny $\pm$ 15.9} &  24.5 \text{\tiny $\pm$ 7.5} \\
         & 80   &   0.0 \text{\tiny $\pm$ 0.0} &   0.0 \text{\tiny $\pm$ 0.0} &   2.0 \text{\tiny $\pm$ 1.2} &   0.0 \text{\tiny $\pm$ 0.1} &    0.6 \text{\tiny $\pm$ 0.8} &   0.0 \text{\tiny $\pm$ 0.0} \\
         & 160  &   0.0 \text{\tiny $\pm$ 0.0} &   0.0 \text{\tiny $\pm$ 0.0} &   0.0 \text{\tiny $\pm$ 0.0} &   0.0 \text{\tiny $\pm$ 0.0} &    0.0 \text{\tiny $\pm$ 0.0} &   0.0 \text{\tiny $\pm$ 0.0} \\
         & 800  &   0.0 \text{\tiny $\pm$ 0.0} &   0.0 \text{\tiny $\pm$ 0.0} &   0.0 \text{\tiny $\pm$ 0.0} &   0.0 \text{\tiny $\pm$ 0.0} &    0.0 \text{\tiny $\pm$ 0.0} &   0.0 \text{\tiny $\pm$ 0.0} \\
         & 1600 &   0.0 \text{\tiny $\pm$ 0.0} &   0.0 \text{\tiny $\pm$ 0.0} &   0.0 \text{\tiny $\pm$ 0.0} &   0.0 \text{\tiny $\pm$ 0.0} &    0.0 \text{\tiny $\pm$ 0.0} &   0.0 \text{\tiny $\pm$ 0.0} \\
\midrule
WS & 16   &  30.0 \text{\tiny $\pm$ 4.1} &  20.4 \text{\tiny $\pm$ 5.0} &  73.2 \text{\tiny $\pm$ 9.1} &  65.7 \text{\tiny $\pm$ 8.8} &  44.4 \text{\tiny $\pm$ 18.0} &  14.8 \text{\tiny $\pm$ 5.4} \\
         & 80   &   0.0 \text{\tiny $\pm$ 0.0} &   0.0 \text{\tiny $\pm$ 0.0} &   0.3 \text{\tiny $\pm$ 0.3} &   0.0 \text{\tiny $\pm$ 0.0} &    0.1 \text{\tiny $\pm$ 0.1} &   0.0 \text{\tiny $\pm$ 0.0} \\
\bottomrule
\end{tabular}

%% file: results/algorithmwise/mst_prim_node_acc.tex
\begin{tabular}{llllllll}
\toprule
         &      &                         GINE &                     GINE (H) &                          PGN &                      PGN (H) &                        RecGNN &                   RecGNN (H) \\
\midrule
ER & 16   &  92.6 \text{\tiny $\pm$ 0.8} &  89.6 \text{\tiny $\pm$ 1.7} &  97.3 \text{\tiny $\pm$ 0.4} &  96.4 \text{\tiny $\pm$ 0.6} &   94.2 \text{\tiny $\pm$ 2.3} &  87.5 \text{\tiny $\pm$ 2.4} \\
         & 80   &  79.1 \text{\tiny $\pm$ 1.3} &  51.6 \text{\tiny $\pm$ 4.5} &  89.1 \text{\tiny $\pm$ 1.6} &  79.7 \text{\tiny $\pm$ 3.8} &  70.7 \text{\tiny $\pm$ 27.8} &  29.0 \text{\tiny $\pm$ 6.7} \\
         & 160  &  77.6 \text{\tiny $\pm$ 1.7} &  49.5 \text{\tiny $\pm$ 4.3} &  84.6 \text{\tiny $\pm$ 1.7} &  75.6 \text{\tiny $\pm$ 4.5} &  66.6 \text{\tiny $\pm$ 28.2} &  25.7 \text{\tiny $\pm$ 6.6} \\
         & 800  &  74.5 \text{\tiny $\pm$ 2.0} &  45.0 \text{\tiny $\pm$ 4.2} &  75.7 \text{\tiny $\pm$ 2.0} &  69.5 \text{\tiny $\pm$ 5.5} &  58.9 \text{\tiny $\pm$ 29.0} &  21.3 \text{\tiny $\pm$ 6.4} \\
         & 1600 &  72.9 \text{\tiny $\pm$ 2.2} &  43.2 \text{\tiny $\pm$ 4.0} &  71.9 \text{\tiny $\pm$ 2.1} &  66.8 \text{\tiny $\pm$ 5.1} &  56.0 \text{\tiny $\pm$ 28.5} &  20.1 \text{\tiny $\pm$ 6.3} \\
\midrule
WS & 16   &  89.6 \text{\tiny $\pm$ 1.4} &  86.0 \text{\tiny $\pm$ 2.1} &  96.8 \text{\tiny $\pm$ 1.0} &  96.1 \text{\tiny $\pm$ 1.0} &   92.8 \text{\tiny $\pm$ 2.8} &  82.0 \text{\tiny $\pm$ 4.0} \\
         & 80   &  75.3 \text{\tiny $\pm$ 1.0} &  54.9 \text{\tiny $\pm$ 6.2} &  82.5 \text{\tiny $\pm$ 2.4} &  74.5 \text{\tiny $\pm$ 3.9} &  67.4 \text{\tiny $\pm$ 22.9} &  32.0 \text{\tiny $\pm$ 7.3} \\
\bottomrule
\end{tabular}

%% file: results/algorithmwise/fast_mis_graph_accuracy.tex
\begin{tabular}{llllllll}
\toprule
         &      &                         GIN &                     GIN (H) &                           PGN &                      PGN (H) &                        RecGNN &                   RecGNN (H) \\
\midrule
ER & 16   &  6.2 \text{\tiny $\pm$ 3.2} &  3.3 \text{\tiny $\pm$ 2.5} &   98.8 \text{\tiny $\pm$ 0.2} &  98.6 \text{\tiny $\pm$ 0.4} &  56.1 \text{\tiny $\pm$ 13.1} &  44.1 \text{\tiny $\pm$ 5.8} \\
         & 80   &  0.0 \text{\tiny $\pm$ 0.0} &  0.0 \text{\tiny $\pm$ 0.0} &   89.2 \text{\tiny $\pm$ 4.6} &  88.9 \text{\tiny $\pm$ 3.1} &    5.5 \text{\tiny $\pm$ 7.1} &   2.6 \text{\tiny $\pm$ 1.5} \\
         & 160  &  0.0 \text{\tiny $\pm$ 0.0} &  0.0 \text{\tiny $\pm$ 0.0} &  74.1 \text{\tiny $\pm$ 10.1} &  76.2 \text{\tiny $\pm$ 7.3} &    0.8 \text{\tiny $\pm$ 1.6} &   0.1 \text{\tiny $\pm$ 0.2} \\
         & 800  &  0.0 \text{\tiny $\pm$ 0.0} &  0.0 \text{\tiny $\pm$ 0.0} &  10.7 \text{\tiny $\pm$ 10.5} &  18.0 \text{\tiny $\pm$ 8.6} &    0.0 \text{\tiny $\pm$ 0.0} &   0.0 \text{\tiny $\pm$ 0.0} \\
         & 1600 &  0.0 \text{\tiny $\pm$ 0.0} &  0.0 \text{\tiny $\pm$ 0.0} &    2.0 \text{\tiny $\pm$ 2.5} &   5.2 \text{\tiny $\pm$ 4.3} &    0.0 \text{\tiny $\pm$ 0.0} &   0.0 \text{\tiny $\pm$ 0.0} \\
\midrule
WS & 16   &  6.5 \text{\tiny $\pm$ 2.8} &  4.4 \text{\tiny $\pm$ 2.3} &   98.1 \text{\tiny $\pm$ 0.6} &  98.2 \text{\tiny $\pm$ 0.3} &  52.6 \text{\tiny $\pm$ 14.6} &  46.5 \text{\tiny $\pm$ 5.7} \\
         & 80   &  0.0 \text{\tiny $\pm$ 0.0} &  0.0 \text{\tiny $\pm$ 0.0} &   84.4 \text{\tiny $\pm$ 8.4} &  82.2 \text{\tiny $\pm$ 7.4} &    9.0 \text{\tiny $\pm$ 9.8} &   4.2 \text{\tiny $\pm$ 1.1} \\
\bottomrule
\end{tabular}

%% file: results/algorithmwise/fast_mis_node_acc.tex
\begin{tabular}{llllllll}
\toprule
         &      &                          GIN &                      GIN (H) &                          PGN &                      PGN (H) &                       RecGNN &                   RecGNN (H) \\
\midrule
ER & 16   &  82.2 \text{\tiny $\pm$ 2.5} &  79.9 \text{\tiny $\pm$ 2.9} &  99.8 \text{\tiny $\pm$ 0.1} &  99.8 \text{\tiny $\pm$ 0.1} &  93.6 \text{\tiny $\pm$ 2.2} &  92.2 \text{\tiny $\pm$ 0.7} \\
         & 80   &  81.6 \text{\tiny $\pm$ 1.9} &  79.9 \text{\tiny $\pm$ 2.2} &  99.6 \text{\tiny $\pm$ 0.2} &  99.4 \text{\tiny $\pm$ 0.1} &  90.0 \text{\tiny $\pm$ 2.3} &  88.7 \text{\tiny $\pm$ 1.7} \\
         & 160  &  80.8 \text{\tiny $\pm$ 2.4} &  78.2 \text{\tiny $\pm$ 2.7} &  99.5 \text{\tiny $\pm$ 0.2} &  99.4 \text{\tiny $\pm$ 0.2} &  90.1 \text{\tiny $\pm$ 2.5} &  88.3 \text{\tiny $\pm$ 2.8} \\
         & 800  &  83.6 \text{\tiny $\pm$ 1.5} &  83.4 \text{\tiny $\pm$ 0.8} &  98.8 \text{\tiny $\pm$ 0.6} &  98.8 \text{\tiny $\pm$ 0.5} &  87.9 \text{\tiny $\pm$ 1.9} &  85.6 \text{\tiny $\pm$ 3.5} \\
         & 1600 &  80.8 \text{\tiny $\pm$ 2.5} &  79.2 \text{\tiny $\pm$ 1.6} &  98.9 \text{\tiny $\pm$ 0.5} &  98.9 \text{\tiny $\pm$ 0.7} &  88.2 \text{\tiny $\pm$ 2.6} &  84.7 \text{\tiny $\pm$ 5.5} \\
\midrule
WS & 16   &  84.2 \text{\tiny $\pm$ 2.1} &  83.1 \text{\tiny $\pm$ 1.9} &  99.8 \text{\tiny $\pm$ 0.1} &  99.7 \text{\tiny $\pm$ 0.1} &  93.3 \text{\tiny $\pm$ 2.2} &  92.9 \text{\tiny $\pm$ 0.5} \\
         & 80   &  82.0 \text{\tiny $\pm$ 2.6} &  79.5 \text{\tiny $\pm$ 3.4} &  99.4 \text{\tiny $\pm$ 0.3} &  99.5 \text{\tiny $\pm$ 0.2} &  92.6 \text{\tiny $\pm$ 2.6} &  92.2 \text{\tiny $\pm$ 1.9} \\
\bottomrule
\end{tabular}

%% file: results/algorithmwise/fast_mis_node_f1.tex
\begin{tabular}{llllllll}
\toprule
         &      &                           GIN &                       GIN (H) &                          PGN &                      PGN (H) &                        RecGNN &                    RecGNN (H) \\
\midrule
ER & 16   &   61.2 \text{\tiny $\pm$ 7.0} &  52.1 \text{\tiny $\pm$ 14.2} &  99.6 \text{\tiny $\pm$ 0.1} &  99.6 \text{\tiny $\pm$ 0.1} &   87.7 \text{\tiny $\pm$ 4.3} &   84.9 \text{\tiny $\pm$ 1.0} \\
         & 80   &  48.2 \text{\tiny $\pm$ 11.1} &  26.8 \text{\tiny $\pm$ 17.7} &  99.1 \text{\tiny $\pm$ 0.4} &  98.8 \text{\tiny $\pm$ 0.3} &   76.5 \text{\tiny $\pm$ 6.8} &   73.5 \text{\tiny $\pm$ 5.2} \\
         & 160  &  51.7 \text{\tiny $\pm$ 11.1} &  29.4 \text{\tiny $\pm$ 18.1} &  98.9 \text{\tiny $\pm$ 0.5} &  98.7 \text{\tiny $\pm$ 0.5} &   78.5 \text{\tiny $\pm$ 6.6} &   74.4 \text{\tiny $\pm$ 8.0} \\
         & 800  &  29.5 \text{\tiny $\pm$ 17.2} &    9.5 \text{\tiny $\pm$ 8.1} &  96.4 \text{\tiny $\pm$ 1.8} &  96.4 \text{\tiny $\pm$ 1.6} &  61.8 \text{\tiny $\pm$ 10.9} &  51.8 \text{\tiny $\pm$ 19.0} \\
         & 1600 &  41.1 \text{\tiny $\pm$ 14.6} &  18.3 \text{\tiny $\pm$ 13.7} &  97.3 \text{\tiny $\pm$ 1.3} &  97.4 \text{\tiny $\pm$ 1.7} &   70.6 \text{\tiny $\pm$ 9.4} &  60.1 \text{\tiny $\pm$ 19.3} \\
\midrule
WS & 16   &   57.5 \text{\tiny $\pm$ 7.2} &  50.8 \text{\tiny $\pm$ 13.1} &  99.4 \text{\tiny $\pm$ 0.3} &  99.3 \text{\tiny $\pm$ 0.2} &   84.0 \text{\tiny $\pm$ 5.0} &   82.4 \text{\tiny $\pm$ 1.1} \\
         & 80   &   63.5 \text{\tiny $\pm$ 7.1} &  50.5 \text{\tiny $\pm$ 17.4} &  99.0 \text{\tiny $\pm$ 0.6} &  99.0 \text{\tiny $\pm$ 0.4} &   85.6 \text{\tiny $\pm$ 5.3} &   85.2 \text{\tiny $\pm$ 3.5} \\
\bottomrule
\end{tabular}

%% file: results/algorithmwise/eccentricity_graph_accuracy.tex
\resizebox{1.0\textwidth}{!}{
\begin{tabular}{llllllll}
\toprule
         &      &                           GIN &                       GIN (H) &                           PGN &                       PGN (H) &                        RecGNN &                    RecGNN (H) \\
\midrule
ER & 16   &  57.3 \text{\tiny $\pm$ 21.2} &  25.3 \text{\tiny $\pm$ 41.0} &  100.0 \text{\tiny $\pm$ 0.0} &  100.0 \text{\tiny $\pm$ 0.0} &  75.8 \text{\tiny $\pm$ 26.2} &   95.0 \text{\tiny $\pm$ 6.3} \\
         & 80   &  77.1 \text{\tiny $\pm$ 17.5} &  23.8 \text{\tiny $\pm$ 39.0} &  100.0 \text{\tiny $\pm$ 0.0} &  100.0 \text{\tiny $\pm$ 0.0} &  80.5 \text{\tiny $\pm$ 35.0} &   96.6 \text{\tiny $\pm$ 3.6} \\
         & 160  &  72.3 \text{\tiny $\pm$ 18.0} &  26.1 \text{\tiny $\pm$ 36.8} &  100.0 \text{\tiny $\pm$ 0.0} &  100.0 \text{\tiny $\pm$ 0.0} &  75.0 \text{\tiny $\pm$ 39.1} &   95.8 \text{\tiny $\pm$ 4.6} \\
         & 800  &  51.3 \text{\tiny $\pm$ 34.2} &  17.1 \text{\tiny $\pm$ 32.9} &  100.0 \text{\tiny $\pm$ 0.0} &  100.0 \text{\tiny $\pm$ 0.0} &  72.7 \text{\tiny $\pm$ 27.9} &  93.4 \text{\tiny $\pm$ 10.3} \\
         & 1600 &  36.7 \text{\tiny $\pm$ 17.6} &  16.0 \text{\tiny $\pm$ 21.7} &  64.6 \text{\tiny $\pm$ 14.9} &   83.0 \text{\tiny $\pm$ 6.5} &  63.0 \text{\tiny $\pm$ 24.8} &  72.1 \text{\tiny $\pm$ 20.9} \\
\midrule
WS & 16   &  78.0 \text{\tiny $\pm$ 18.7} &  25.3 \text{\tiny $\pm$ 42.2} &  100.0 \text{\tiny $\pm$ 0.0} &  100.0 \text{\tiny $\pm$ 0.0} &  86.7 \text{\tiny $\pm$ 25.7} &   99.0 \text{\tiny $\pm$ 1.2} \\
         & 80   &  27.6 \text{\tiny $\pm$ 19.5} &  19.0 \text{\tiny $\pm$ 18.8} &   93.8 \text{\tiny $\pm$ 2.1} &   88.3 \text{\tiny $\pm$ 1.8} &  60.8 \text{\tiny $\pm$ 29.1} &  66.4 \text{\tiny $\pm$ 22.4} \\
\bottomrule
\end{tabular}}

%% file: results/algorithmwise/eccentricity_mse.tex
\resizebox{1.0\textwidth}{!}{
\begin{tabular}{llllllll}
\toprule
         &      &                                 GIN &                             GIN (H) &                                PGN &                            PGN (H) &                             RecGNN &                         RecGNN (H) \\
\midrule
ER & 16   &        47.7 \text{\tiny $\pm$ 27.6} &      186.8 \text{\tiny $\pm$ 154.6} &         0.0 \text{\tiny $\pm$ 0.0} &         0.0 \text{\tiny $\pm$ 0.0} &        14.8 \text{\tiny $\pm$ 9.6} &         6.2 \text{\tiny $\pm$ 5.2} \\
         & 80   &        26.0 \text{\tiny $\pm$ 20.5} &      248.5 \text{\tiny $\pm$ 192.5} &         0.0 \text{\tiny $\pm$ 0.0} &         0.0 \text{\tiny $\pm$ 0.0} &       11.1 \text{\tiny $\pm$ 11.8} &         5.2 \text{\tiny $\pm$ 3.7} \\
         & 160  &        43.9 \text{\tiny $\pm$ 34.2} &      274.6 \text{\tiny $\pm$ 211.9} &         0.1 \text{\tiny $\pm$ 0.1} &         0.1 \text{\tiny $\pm$ 0.0} &       13.7 \text{\tiny $\pm$ 11.5} &         5.0 \text{\tiny $\pm$ 3.6} \\
         & 800  &        47.3 \text{\tiny $\pm$ 40.1} &      442.7 \text{\tiny $\pm$ 390.5} &         1.4 \text{\tiny $\pm$ 2.3} &         0.2 \text{\tiny $\pm$ 0.0} &       15.0 \text{\tiny $\pm$ 11.2} &         6.3 \text{\tiny $\pm$ 5.6} \\
         & 1600 &        93.4 \text{\tiny $\pm$ 43.7} &      357.5 \text{\tiny $\pm$ 210.3} &        22.7 \text{\tiny $\pm$ 5.9} &        12.2 \text{\tiny $\pm$ 0.8} &       33.7 \text{\tiny $\pm$ 13.9} &       29.0 \text{\tiny $\pm$ 21.9} \\
\midrule
WS & 16   &        18.8 \text{\tiny $\pm$ 11.0} &      220.6 \text{\tiny $\pm$ 181.6} &         0.0 \text{\tiny $\pm$ 0.0} &         0.0 \text{\tiny $\pm$ 0.0} &        10.2 \text{\tiny $\pm$ 9.4} &         2.9 \text{\tiny $\pm$ 2.2} \\
         & 80   &      501.0 \text{\tiny $\pm$ 301.6} &      524.5 \text{\tiny $\pm$ 277.9} &     153.3 \text{\tiny $\pm$ 102.0} &       87.7 \text{\tiny $\pm$ 28.9} &       98.9 \text{\tiny $\pm$ 64.1} &     306.6 \text{\tiny $\pm$ 207.0} \\
\bottomrule
\end{tabular}
}